\documentclass[10pt,journal,compsoc]{IEEEtran}
\ifCLASSOPTIONcompsoc
  \usepackage[nocompress]{cite}
\else
  \usepackage{cite}
\fi
\ifCLASSINFOpdf
\else
\fi
\usepackage{amsmath,amsthm}
\usepackage{amssymb}
\usepackage{bbm}
\usepackage{amsfonts}
\usepackage{graphicx}
\usepackage{subfigure}
\graphicspath{{graphics/}}
\usepackage{tabularx,booktabs}
\usepackage{tikz}
\usepackage[ruled,vlined,linesnumbered]{algorithm2e}
\usepackage{algorithmic}

\usepackage{xcolor}
\usepackage{hyperref}
\hypersetup{
    colorlinks=true,
    linkcolor=blue,
    filecolor=magenta,      
    urlcolor=blue,
    citecolor=blue,
}
\usepackage{booktabs,caption}
\usepackage{multirow}
\usepackage{lscape}
\usepackage{epstopdf}
\usepackage{lineno}
\usepackage{subfigure}

\newcommand{\given}{\;\middle|\;}

\newcommand{\bd}[1]{\boldsymbol{#1}}

\begin{document}
%

\title{Bayesian Complementary Kernelized Learning for Multidimensional Spatiotemporal Data}

\author{Mengying~Lei,
        Aur\'elie~Labbe,
        and~Lijun~Sun\textsuperscript{$\ast$},~\IEEEmembership{Senior Member,~IEEE}
\IEEEcompsocitemizethanks{\IEEEcompsocthanksitem Mengying Lei and Lijun Sun are with the Department of Civil Engineering, McGill University, Montreal, Quebec, H3A 0C3, Canada. E-mail: mengying.lei@mail.mcgill.ca (Mengying Lei), lijun.sun@mcgill.ca (Lijun Sun).
\IEEEcompsocthanksitem Aur\'elie Labbe is with the Department of Decision Sciences, HEC Montreal, Montreal, Quebec, H3T 2A7, Canada. E-mail: aurelie.labbe@hec.ca.
\IEEEcompsocthanksitem \textsuperscript{$\ast$}Corresponding author.}
\thanks{Manuscript received xx, 2022; revised xx.}}

\IEEEtitleabstractindextext{%
\begin{abstract}
Probabilistic modeling of multidimensional spatiotemporal data is critical to many real-world applications. As real-world spatiotemporal data often exhibits complex dependencies that are nonstationary and nonseparable, developing effective and computationally efficient statistical models to accommodate nonstationary/nonseparable processes containing both long-range and short-scale variations becomes a challenging task, in particular for large-scale datasets with various corruption/missing structures. In this paper, we propose a new statistical framework---Bayesian Complementary Kernelized Learning (BCKL)---to achieve scalable probabilistic modeling for multidimensional spatiotemporal data. To effectively characterize complex dependencies, BCKL integrates two complementary approaches--- kernelized low-rank tensor factorization and short-range spatiotemporal Gaussian Processes.  Specifically, we use a multi-linear low-rank factorization component to capture the global/long-range correlations in the data and introduce an additive short-scale GP based on compactly supported kernel functions to characterize the remaining local variabilities. We develop an efficient Markov chain Monte Carlo (MCMC) algorithm for model inference and evaluate the proposed BCKL framework on both synthetic and real-world spatiotemporal datasets. Our experiment results show that BCKL offers superior performance in providing accurate posterior mean and high-quality uncertainty estimates, confirming the importance of both global and local components in modeling spatiotemporal data.






\end{abstract}


}

\maketitle

\IEEEdisplaynontitleabstractindextext

%
\IEEEpeerreviewmaketitle

\IEEEraisesectionheading{\section{Introduction}\label{sec:introduction}}

\IEEEPARstart{P}robabilistic modeling for  real-world spatiotemporal data is crucial to many scientific fields, such as ecology, environmental sciences, epidemiology, remote sensing, meteorology, and climate science, to name but a few \cite{banerjee2014hierarchical, cressie2015statistics, banerjee2017high}. In many applications with a predefined spatial and temporal domain, the research question can be generalized to learning the variance structure, and performing interpolation on a multidimensional Cartesian product space (or a grid) with both mean and uncertainty estimated. Since real-world datasets are often large-scale, highly sparse, and show complex spatiotemporal dependencies, developing efficient and effective probabilistic spatiotemporal models becomes a significant challenge in geostatistics and machine learning.

Essentially, there exist two key solutions for statistical spatiotemporal modeling on a grid with uncertainty quantification: Bayesian multi-linear matrix/tensor factorization \cite{salakhutdinov2008bayesian,chen2021bayesian,lei2022bayesian} and hierarchical multivariate/multidimensional (i.e., vector/matrix/tensor-valued) Gaussian process (GP) regression \cite{bonilla2007multi, alvarez2012kernels, borchani2015survey}. Given incomplete two-dimensional (2D) spatiotemporal data $\boldsymbol{Y}\in\mathbb{R}^{M\times N}$ defined on $M$ locations over $N$ time points and partially observed on an index set $\Omega$ ($|\Omega|<MN$), matrix factorization (MF) assumes that $\boldsymbol{Y}$ can be characterized by a low-rank structure along with independently distributed residual/noise: $\boldsymbol{Y}=\boldsymbol{U}\boldsymbol{V}^{\top}+\boldsymbol{E}$, where $\boldsymbol{U}\in\mathbb{R}^{M\times D}$, $\boldsymbol{V}\in\mathbb{R}^{N\times D}$ are the latent spatial and temporal factor matrices with rank $D\ll \{M,N\}$, and $\boldsymbol{E}\in\mathbb{R}^{M\times N}$ is a noise matrix with each entry following $\epsilon_{mn}\sim\mathcal{N}\left(0,\tau_{mn}^{-1}\right)$. One can also make different assumptions on the noise level, such as $\tau_{mn}=\tau$ in probabilistic principle component analysis (PCA) and $\tau_{mn}=\tau_m$ or $\tau_{mn}=\tau_n$ in factor analysis (FA). Low-rank factorization methods offer a natural solution to model sparse data with a large number of missing values, and representative applications of MF include image inpainting \cite{zhao2015bayesian}, collaborative filtering \cite{salakhutdinov2008bayesian,xiong2010temporal}, and spatiotemporal data (e.g., traffic speed/flow) imputation \cite{chen2021bayesian, lei2022bayesian}.  GP regression assumes that  $\boldsymbol{y}=\boldsymbol{f}+\boldsymbol{\epsilon}$, where  $\boldsymbol{y}$ denotes the vectorized version of $\boldsymbol{Y}$, $\boldsymbol{f}$ is a GP with a parametric kernel function, and $\boldsymbol{\epsilon}$ denotes Gaussian noises with precision $\tau$. 
Thanks to the elegant mathematical properties of Gaussian distributions (e.g., analytical conditional and marginal distributions for prediction and uncertainty quantification), GP has become the primary tool to model diverse types of spatiotemporal phenomena  \cite{banerjee2014hierarchical,gelfand2003spatial}. Both approaches can be extended to a multidimensional setting. For example, for a multivariate spatiotemporal tensor $\boldsymbol{\mathcal{Y}}\in \mathbb{R}^{M\times N\times P}$ with $P$ variables, one can apply either tensor factorization \cite{kolda2009tensor} or the linear model of coregionalization (LMC) \cite{alvarez2012kernels} (or multi-task GP) to model the data, by introducing new factor matrices and new kernel structures, respectively.

The first low-rank factorization framework is built, by default, based solely on linear algebra with no requirement for spatiotemporal information (e.g., the distance between two locations). Therefore, the default models are invariant to permutations in rows and columns (i.e., space and time), and the results lack spatiotemporal consistency/smoothness. To address this issue, recent work has introduced different smoothness constraints such as graph Laplacian regularization for encoding spatial correlations \cite{yokota2016smooth,bahadori2014fast, rao2015collaborative}, time series models for encoding temporal evolution \cite{gamerman2008spatial, xiong2010temporal, chen2021bayesian}, and flexible GP priors on both spatial and temporal factors with well-designed kernel functions \cite{luttinen2009variational, lei2022bayesian}. For the GP regression approach, a critical challenge is the cubic computational cost of model inference, and the fundamental research question is to design computationally efficient covariance structures to characterize complex spatiotemporal dependencies. One commonly used approach is through a separable kernel formed by the product of stationary covariance functions defined on each input dimension, which results in a covariance matrix with a Kronecker product structure that can be leveraged for scalable GP inference \cite{saatcci2012scalable}. However, the computational advantage brought by the Kronecker product disappears when the data contains missing values, and more importantly, the separable structure 
has limited capacity to model real-world spatiotemporal processes that are often nonseparable. Several studies in the machine learning community have tried to develop expressive GP models for incomplete large-scale multidimensional data, for example, based on spectral mixture kernels \cite{wilson2014fast, wilson2015kernel, remes2017non}. For such methods, computing the uncertainty for the entire grid is intractable, and the covariance could be difficult to interpret. An alternative strategy for computationally manageable GP modeling is to introduce sparsity into the covariance matrix using compactly supported kernels, such as in \cite{furrer2006covariance, luttinen2012efficient}. These models are effective in characterizing short-scale local variations, but long-range correlations are explicitly ignored due to  restrictions of the sparse covariance.

Although both solutions are capable of modeling large-scale incomplete spatiotemporal data, they are designed following different assumptions and serve different applications. The low-rank factorization model focuses on explaining the global structure using a few latent factors, while GP regression characterizes local correlations through kernel functions determined by a few hyperparameters. Real-world datasets, however, often exhibit more complicated patterns with both long-range and nonstationary global structures and short-scale local variations. For example, in traffic speed/flow data, there exist both global daily/weekly periodic patterns due to strong regularity in human travel behavior and short-period local perturbations caused by traffic incidents and other events \cite{li2015trend}. To accurately model such data, the low-rank framework will require many factors to accommodate those local perturbations. One will likely observe spatiotemporally correlated residuals when using a small rank. While for the second GP model, the commonly used stationary and separable covariance structure has minimal capacity to encode the nonstationary and nonseparable correlations and account for the global/long-range patterns beyond periodicity. {On the temporal dimension, although it is possible to design complex covariance structures to encode global/periodic structures by combining periodic kernels with other kernels through sum/product operations, we still suffer from kernel hyperparameter identifiability issues and the prohibitive computational cost \cite{wang2020non}. }


In this paper, we propose a Bayesian Complementary Kernelized Learning (BCKL) framework which integrates the two  solutions in a single model. The key idea is to fit the spatiotemporal data with a GP where the mean function is parameterized by low-rank factorization and the covariance matrix is formed by a sum of separable product kernels with compact support. In doing so, the nonstationary and nonseparable dependencies can be effectively explained through the combination of two complementary modules: low-rank factorization for global patterns and short-scale GP  for local variations. Although similar ideas have been developed for large-scale spatial data by combining predictive processes with covariance tapering \cite{sang2012full}, BCKL represents a novel approach that utilizes tensor factorization to model the global component for multidimensional/spatiotemporal processes.  BCKL is a scalable framework that inherits the high computational efficiency of both low-rank factorization and short-scale GP, and can be viewed from two different perspectives as i) a generalized probabilistic low-rank factorization model with spatiotemporally correlated errors, and ii) a local GP regression with the mean function characterized by a latent factor model. The main contribution of this work is threefold:
\begin{enumerate}
    \item We combine kernelized low-rank factorization and short-scale GP (via covariance tapering) into an integrated framework to efficiently and effectively  model spatiotemporal data. The low-rank matrix or tensor factorization can leverage the structural correlation among different input dimensions/modes (e.g., space, time, and variables), thus providing an interpretable and highly efficient solution to capture nonstationary global trends and long-range dependencies in the data. 
    \item The proposed model is fully Bayesian and we derive an efficient Markov chain Monte Carlo (MCMC) algorithm, which decouples the inference of the two components by explicitly sampling the local GP as latent variables. The Bayesian framework provides posterior distributions for accurate data estimation with high-quality uncertainty quantification, which is important for many risk-sensitive applications and decision-making processes. 
    \item We conduct extensive experiments on both synthetic and real-world spatiotemporal datasets to demonstrate the effectiveness of BCKL.
\end{enumerate}

The remainder of this paper is organized as follows. In Section~\ref{sec:RelatedWork}, related studies for low-rank factorization and GP modeling are briefly reviewed. In Section~\ref{sec:preliminary}, we introduce basic model formulations for Bayesian kernelized low-rank factorization and GP regression. In Section~\ref{sec:method}, we present the specification of BCKL and the sampling algorithm for model inference. In Section~\ref{sec:experiments}, we conduct comprehensive experiments on synthetic and real-world spatiotemporal data. Section~\ref{sec:conclusion} summarizes this work and discusses future research directions.


\section{Related Work}
\label{sec:RelatedWork}

Our key research question is to develop efficient and effective probabilistic models for large-scale and multidimensional spatiotemporal data with complex nonstationary and nonseparable correlations. As mentioned, there are two general solutions for this task: low-rank matrix/tensor factorization and multivariate (e.g., vector-valued/matrix-valued) GP. In this section, we mainly review related studies and also introduce some work that develops complementary global/local kernels in other domains.

Low-rank matrix/tensor factorization is a widely used approach for modeling multidimensional datasets with missing entries. Spatial/temporal priors can be introduced on the lower-dimensional latent factors (see e.g., \cite{bahadori2014fast, rao2015collaborative, yu2016temporal} with graph and autoregressive regularization) to enhance model performance. However, this approach is often formulated as an optimization problem, which requires extensive tuning of regularization parameters to achieve optimal results.  In contrast, Bayesian kernelized low-rank models provide a more consistent framework that learns the posterior distributions for kernel hyperparameters and data with  uncertainty quantification. Examples of kernelized low-rank models include the spatial factor model \cite{gamerman2008spatial,ren2013hierarchical} and Bayesian GP factorization   \cite{luttinen2009variational,lei2022bayesian,lei2021scalable}. Although these spatiotemporal low-rank factorization methods are effective in modeling the global structure of the data, they tend to generate over-smoothed structures when applied to real-world spatiotemporal datasets. For instance, in the aforementioned traffic data case, a large number of factors may be required to effectively capture the many small-scale variations, resulting in increased computational costs. 

The second solution is to model  multivariate and multidimensional spatiotemporal data directly using Gaussian processes. However, given the large size of the data (e.g., $|\Omega|\propto M\times N\times P$) and the cubic time complexity for standard GP, the critical question is to design efficient kernel structures to characterize complex relationships within the data. For this purpose, various kernel configurations have been developed in the literature. The well-known LMC (linear model of coregionalization) for multivariate spatial processes (see \cite{schmidt2003bayesian, gelfand2004nonstationary,bonilla2007multi} for some examples and \cite{banerjee2014hierarchical,alvarez2012kernels} for a summary) provides a general construction form for multivariate and multidimensional problems. In addition, a widely used configuration is the separable product kernel, based on which exact inference for multidimensional data can be achieved by leveraging the Kronecker product structure of the covariance matrix with substantially reduced time cost \cite{saatcci2012scalable, wilson2014fast}. However, such a simplified kernel structure has limited capacity in modeling complex (e.g., nonstationary and nonseparable) multivariate spatiotemporal processes. 
Another possible solution is to build sparse covariance matrices using compactly supported kernels, such as  covariance tapering \cite{furrer2006covariance}. Nevertheless, due to the sparse nature of the kernel function, these models can only capture short-scale variations and fail to encode long-range dependencies. Another key issue in the GP-based approach is that most models make a zero-mean assumption and focus only on modeling the covariance matrix, while the importance of a proper mean function is often overlooked. However, the mean structure could play an important role in interpolation and extrapolation \cite{heaton2019case}. Some mean functions for GP modeling have been discussed in \cite{kaufman2011efficient, gu2022gaussian}, but they typically assume a naive constant or polynomial regression surface.

In this paper, we propose a complementary framework that combines a global low-rank factorization model and a local multidimensional GP. The most related studies are kernelized matrix factorization \cite{luttinen2009variational,lei2022bayesian} to build the global component and spatiotemporal nonseparable short-scale GP \cite{luttinen2012efficient} to build the local process, both in a 2D matrix setting. The combination of low-rank and sparse matrices is also introduced in \cite{sang2012full} for covariance approximation, but only for spatial data. Compared with prior studies, the proposed method can be considered as building a multidimensional spatiotemporal process in which the mean structure is modeled by kernelized low-rank factorization, which is similar to probabilistic Karhunen–Lo\`{e}ve expansion and functional principle component analysis (FPCA) \cite{wang2008karhunen}. With this model assumption, both long-range global trends and short-scale local variations of data can be effectively characterized. The complementary idea is also closely related to the recent work \cite{descary2019functional, masak2022random} on functional data analysis, which solves  covariance estimation as an optimization problem. However, our work is fundamentally different as we pursue a fully Bayesian statistical model with the capacity for uncertainty quantification.

\section{Preliminaries}\label{sec:preliminary}

Throughout this paper, we use lowercase letters to denote scalars, e.g., $x$, boldface lowercase letters to denote vectors, e.g., $\boldsymbol{x}\in\mathbb{R}^{M}$, and boldface uppercase letters to denote matrices, e.g., $\boldsymbol{X}\in\mathbb{R}^{M\times N}$. The $\ell_{2}$-norm of $\boldsymbol{x}$ is defined as $\|\boldsymbol{x}\|_{2}=\sqrt{\sum_{m}x_{m}^{2}}$. For a matrix $\boldsymbol{X}\in\mathbb{R}^{M\times N}$, we denote its $(m,n)$th entry by $x_{mn}$ or $\boldsymbol{X}(m,n)$, and its determinant by $\left|\boldsymbol{X}\right|$. We use $\bd{I}$ or $\bd{I}_{N}$ to represent an identity matrix of size $N$. Given two matrices $\boldsymbol{A}\in\mathbb{R}^{M\times N}$ and $\boldsymbol{B}\in\mathbb{R}^{P\times Q}$, the Kronecker product is defined as $\boldsymbol{A}\otimes\boldsymbol{B}=
\footnotesize{\begin{bmatrix}
  a_{1,1}\boldsymbol{B} & \cdots & a_{1,N}\boldsymbol{B} \\
  \vdots & \ddots & \vdots \\
  a_{M,1}\boldsymbol{B} & \cdots & a_{M,N}\boldsymbol{B}
\end{bmatrix}}\in\mathbb{R}^{MP\times NQ}$. The hadamard product, i.e., element-wise product, for two commensurate matrices $\boldsymbol{A}$ and $\boldsymbol{B}$ is denoted by $\boldsymbol{A}\circledast\boldsymbol{B}$, and the outer product of two vectors $\boldsymbol{a}$ and $\boldsymbol{b}$ is denoted by $\boldsymbol{a}\circ\boldsymbol{b}$. The vectorization $\operatorname{vec}(\bd{X})$ stacks all column vectors in $\bd{X}$ as a single vector. Following the tensor notation in \cite{kolda2009tensor}, we denote a third-order tensor by $\boldsymbol{\mathcal{X}}\in\mathbb{R}^{M\times N\times P}$ and its mode-$k$ $(k=\text{1},\text{2},\text{3})$ unfolding by $\boldsymbol{X}_{(k)}$, which maps a tensor into a matrix. The $(m,n,p)$th element of $\boldsymbol{\mathcal{X}}$ is denoted by $x_{mnp}$ or $\boldsymbol{\mathcal{X}}(m,n,p)$, and the vectorization of $\boldsymbol{\mathcal{X}}$ is defined by $\operatorname{vec}(\boldsymbol{\mathcal{X}})=\operatorname{vec}\left(\boldsymbol{X}_{(1)}\right)$.

\subsection{Bayesian kernelized low-rank factorization}

In kernelized low-rank MF \cite{zhou2012kernelized, luttinen2009variational}, each column of the latent factors $\boldsymbol{U}\in\mathbb{R}^{M\times D}$ and $\boldsymbol{V}\in\mathbb{R}^{N\times D}$ is assumed to have a zero mean GP prior. Relevant hyper-priors are further imposed on the kernel hyperparameters and model noise precision $\tau$, respectively, to complete the assumptions for building a Bayesian kernelized MF (BKMF) model \cite{lei2022bayesian}. The generative model of BKMF can be summarized as:
\begin{equation}
\begin{aligned}
&\tau\sim\text{Gamma}\left(a_0,b_0\right), \\  
&\text{for } d=1,2,\ldots,D \\
&\ \ \forall\theta\in\left\{\boldsymbol{\theta}_{u}^{d},\boldsymbol{\theta}_{v}^{d}\right\},\log{\left(\theta\right)}\sim\mathcal{N}\left(\mu_{\theta},\tau_{\theta}^{-1}\right),\\
&\ \ \boldsymbol{u}_{d}\sim\mathcal{N}\left(\boldsymbol{0},\boldsymbol{K}_{u}^{d}\given\boldsymbol{\theta}_u^d\right),~ \boldsymbol{v}_{d}\sim\mathcal{N}\left(\boldsymbol{0},\boldsymbol{K}_{v}^{d}\given\boldsymbol{\theta}_v^d\right), \\
&\text{and }  y_{mn}\sim \mathcal{N}\left(\sum\nolimits_{d=1}^D u_{md} v_{nd},\tau^{-1}\right), \ \forall (m,n)\in \Omega,
\end{aligned}
\end{equation}
where $\boldsymbol{K}_{u}^{d}\in\mathbb{R}^{M\times M}$ and $\boldsymbol{K}_{v}^{d}\in\mathbb{R}^{N\times N}$ are the covariance matrices for the $d$th column in $\boldsymbol{U}$ and $\boldsymbol{V}$, i.e., $\boldsymbol{u}_{d}$ and $\boldsymbol{v}_{d}$, respectively. Hyperparameters (e.g., length-scale and variance) for $\boldsymbol{K}_{u}^{d}$ and $\boldsymbol{K}_{v}^{d}$ are represented by $\boldsymbol{\theta}_u^d$ and $\boldsymbol{\theta}_v^d$, respectively. Both kernel hyperparameters and other model parameters can be efficiently sampled through MCMC. It is straightforward to extend BKMF to higher-order tensor factorization such as in \cite{lei2021scalable}. 

\subsection{Gaussian process (GP) regression}

GP regression has been extensively used by both the machine learning and the statistics communities. Let $\left\{\boldsymbol{x}_{j},y_{j}\right\}_{j=1}^{|\Omega|}$ be a set of input-output data pairs, for each data point: $y=f(\boldsymbol{x})+\epsilon$, where $f(\boldsymbol{x})$ is the function value at location $\boldsymbol{x}$ and $\epsilon$ represents the noise. In GP regression, the prior distribution for $f(\boldsymbol{x})$ is a GP:
\begin{equation}
f(\boldsymbol{x})\sim\mathcal{GP}\left(m(\boldsymbol{x}),k\left(\boldsymbol{x},\boldsymbol{x}'\right)\right),
\end{equation}
where the mean function $m(\boldsymbol{x})$ is generally taken as zero, $\boldsymbol{x}$ and $\boldsymbol{x}'$ are any pair of inputs, and $k\left(\boldsymbol{x},\boldsymbol{x}'\right)$ is a covariance/kernel function. For example, the widely used squared exponential kernel is $k_{SE}\left(\boldsymbol{x},\boldsymbol{x}'\right)=\sigma^2\exp\left(-0.5\| \boldsymbol{x}-\boldsymbol{x}'\|_2^2/l^2\right)$ with $\sigma^2$ and $l$ as the variance and the length-scale hyperparameters. This function is stationary as it only depends on the distance between two points, and thus the kernel function can be simplified to $k_{SE}\left(\boldsymbol{h}\right)$ where $\boldsymbol{h}=\|\boldsymbol{x}-\boldsymbol{x}'\|$. Assuming $\epsilon\sim\mathcal{N}(0,\tau^{-1})$ a white noise process with precision $\tau$, one can write the joint distribution of the observed output values $\boldsymbol{y}=[y_1,\ldots,y_{|\Omega|}]^{\top}$ and the function values at the test locations $\boldsymbol{f}_{\ast}$ under the prior settings, which is a multivariate normal distribution. Deriving the conditional distribution of $\boldsymbol{f}_{\ast}$ yields the predictive distribution for GP regression:
\begin{equation}
\begin{aligned}
\left.\boldsymbol{f}_{\ast}\right|\boldsymbol{y}\sim\mathcal{N}&\Big(\boldsymbol{K}_{\ast}[\boldsymbol{K}+\tau^{-1}\boldsymbol{I}_{|\Omega|}]^{-1}\boldsymbol{y}, \\
    &\boldsymbol{K}_{\ast\ast}-\boldsymbol{K}_{\ast}[\boldsymbol{K}+\tau^{-1}\boldsymbol{I}_{|\Omega|}]^{-1}\boldsymbol{K}_{\ast}^{\top}\Big),
    \end{aligned}
\end{equation}
where $\boldsymbol{K}=\boldsymbol{K}(\boldsymbol{X},\boldsymbol{X})\in\mathbb{R}^{|\Omega|\times |\Omega|}$, $\boldsymbol{K}_{\ast}=\boldsymbol{K}(\boldsymbol{X}_{\ast},\boldsymbol{X})$, and $\boldsymbol{K}_{\ast\ast}=\boldsymbol{K}(\boldsymbol{X}_{\ast},\boldsymbol{X}_{\ast})$ are the covariance matrices between the observed data points, the test and the observed points, and the test points, respectively; $\boldsymbol{X}=[\boldsymbol{x}_{1},\ldots,\boldsymbol{x}_{|\Omega|}]^{\top}$ denotes observed locations and $\boldsymbol{X}_{\ast}$ denotes the locations of the test points. These covariance values are computed using the covariance function $k(\boldsymbol{x},\boldsymbol{x}')$. 
The latent function values $f(\boldsymbol{x})$ can also be analytically marginalized, and the marginal likelihood of $\boldsymbol{y}$ conditioned only on the hyperparameters of the kernel becomes:
\begin{equation} \label{Eq_marginalLikeli}
\begin{aligned}
\log{p\left(\boldsymbol{y}\given\boldsymbol{\theta}\right)}=&-\frac{1}{2}\boldsymbol{y}^{\top}\left(\boldsymbol{K}+\tau^{-1}\boldsymbol{I}_{|\Omega|}\right)^{-1}\boldsymbol{y}\\
    &-\frac{1}{2}\log{\left|\boldsymbol{K}+\tau^{-1}\boldsymbol{I}_{|\Omega|}\right|}-\frac{|\Omega|}{2}\log{2\pi},
\end{aligned}
\end{equation}
where $\boldsymbol{\theta}$ denotes the kernel hyperparameters. Optimizing this log marginal likelihood is the typical approach to learn $\boldsymbol{\theta}$. Due to the calculation of $\left(\boldsymbol{K}+\tau^{-1}\boldsymbol{I}_{|\Omega|}\right)^{-1}$ and $\left|\boldsymbol{K}+\tau^{-1}\boldsymbol{I}_{|\Omega|}\right|$, the computational cost for model inference and prediction is $\mathcal{O}(|\Omega|^3)$, which is the key bottleneck for applying GP regression on large datasets. Note that there also exist kernel functions that can capture temporal periodic dependencies in the data, such as the periodic kernel function $k_{Per}\left(x,x'\right)=k_{Per}\left(h\right)=\sigma^2\exp\left(-2\sin^2(\frac{\pi h}{p}) / l^2\right)$, where $h=\|x-x'\|$ and $p$ represents the period. One can also introduce more flexibility by multiplying the period kernel with a local kernel (e.g., SE). However, we still face the cubic computational cost in training and the fixed period/frequency is often too restricted to model real-world temporal dynamics.

\subsection{Multidimensional GP modeling and covariance tapering}

For a spatiotemporal dataset that is defined on a grid with the input points being a Cartesian product $S\times T=\{(\boldsymbol{s},t):\boldsymbol{s}\in\mathcal{S},t\in\mathcal{T}\}$ where the coordinate sets $\mathcal{S}$ and $\mathcal{T}$ include $M$ and $N$ points, respectively, a common and efficient GP model is to use a separable kernel:
\begin{equation} \label{Eq_productK}
k\left(\left(\boldsymbol{s},t\right),\left(\boldsymbol{s}',t'\right);\boldsymbol{\theta}\right)=k_s\left(\boldsymbol{s},\boldsymbol{s}';\boldsymbol{\theta}_{s}\right)k_t\left(t,t';\boldsymbol{\theta}_t\right),
\end{equation}
where $k_s$ and $k_t$ are covariance functions defined for the spatial and temporal domains, respectively, and $\boldsymbol{\theta}_s$ and $\boldsymbol{\theta}_t$ denote the corresponding kernel hyperparameters. The constructed covariance matrix $\boldsymbol{K}\in\mathbb{R}^{MN\times MN}$ is the Kronecker product of two smaller covariance matrices $\boldsymbol{K}=\boldsymbol{K}_{t}\otimes\boldsymbol{K}_{s}$, where $\boldsymbol{K}_{s}\in\mathbb{R}^{M\times M}$ and $\boldsymbol{K}_{t}\in\mathbb{R}^{N\times N}$ are computed separately over the input spatial and temporal dimension using $k_s$ and $k_t$, respectively. By leveraging the Kronecker product structure, the computational cost for exact kernel hyperparameter learning for a fully observed dataset can be significantly reduced from $\mathcal{O}(M^3N^3)$ to $\mathcal{O}(M^2N+N^2M)$ \cite{saatcci2012scalable}. However, when the data is partially observed on a subset of indices $\Omega$ ($|\Omega|<MN$) (i.e., with missing values), the covariance matrix $\boldsymbol{K}\in\mathbb{R}^{|\Omega|\times |\Omega|}$ no longer possesses Kronecker structure, and the inference becomes very expensive for large datasets.

To achieve fast and scalable GP modeling for incomplete datasets, one strategy is to introduce sparsity into the covariance matrix, such as using compactly supported covariance functions constructed from a covariance tapering  function $k_{taper}(\Delta;\lambda)$ where $\Delta=\|\boldsymbol{x}-\boldsymbol{x}'\|$, which is an isotropic correlation function with a range parameter $\lambda$ \cite{furrer2006covariance,kaufman2008covariance}. The function value of $k_{taper}$ is exactly zero for $\forall \Delta\geq\lambda$. 
Assuming $k_s$ and $k_t$ are both stationary, i.e., $k_s\left(\boldsymbol{s},\boldsymbol{s}';\boldsymbol{\theta}_{s}\right)=k_s\left(\Delta_s;\boldsymbol{\theta}_{s}\right)$ and $k_t\left(t,t';\boldsymbol{\theta}_{t}\right)=k_t\left(\Delta_t;\boldsymbol{\theta}_{t}\right)$ and combining the product kernel in Eq.~\eqref{Eq_productK} and $k_{taper}$, we can obtain the tapered stationary covariance function:
\begin{equation}
\begin{aligned}
k\left(\Delta_s,\Delta_t;\boldsymbol{\theta},\boldsymbol{\lambda}\right)=&\left[k_s\left(\Delta_s;\boldsymbol{\theta}_{s}\right)k_{taper}\left(\Delta_s;\lambda_s\right)\right)] \\
&\left[k_t\left(\Delta_t;\boldsymbol{\theta}_{t}\right)k_{taper}\left(\Delta_t;\lambda_t\right)\right],
\end{aligned}
\end{equation}
where $\Delta_s$ and $\Delta_t$ are the distances between the inputs in space and time, respectively. The tapered covariance matrix becomes $\left(\boldsymbol{K}_{t}\circledast
\boldsymbol{T}_{t}\right)\otimes\left(\boldsymbol{K}_{s}\circledast
\boldsymbol{T}_{s}\right)$, where $\boldsymbol{T}_{s}$ and $\boldsymbol{T}_{t}$ are covariance matrices calculated by $k_{taper}\left(\Delta_s;\lambda_s\right)$ and $k_{taper}\left(\Delta_t;\lambda_t\right)$, respectively. An example of a commonly used tapering function for 2D inputs is the $\text{Wendland}$ taper for 2D inputs \cite{wendland1995piecewise}: $k_{taper}\left(\Delta;\lambda\right)=\left(1-\frac{\Delta}{\lambda}\right)^{4}\left(1+4\frac{\Delta}{\lambda}\right)$ for $\Delta<\lambda$ and equals to zero for $\Delta\geq\lambda$. We can control the degree of sparsity of the covariance matrix $\boldsymbol{K}$ using different $\lambda$. The covariance for noisy observed data, i.e., $\boldsymbol{K}+\tau^{-1}\boldsymbol{I}_{|\Omega|}$, is also sparse. Using sparse matrix algorithms, covariance tapering offers significant computational benefits in model inference. Specifically, to optimize Eq.~\eqref{Eq_marginalLikeli}, one can use sparse Cholesky matrices to compute $\left(\boldsymbol{K}+\tau^{-1}\boldsymbol{I}_{|\Omega|}\right)^{-1}\boldsymbol{y}$ and  $\log{\left|\boldsymbol{K}+\tau^{-1}\boldsymbol{I}_{|\Omega|}\right|}$. It should be noted that the sparse covariance can only encode small-scale variations since long-range correlations are explicitly ignored.

\section{Methodology}\label{sec:method}

\subsection{Model specification}

\begin{figure}[!t]
\centering
\includegraphics[width=0.5\textwidth]{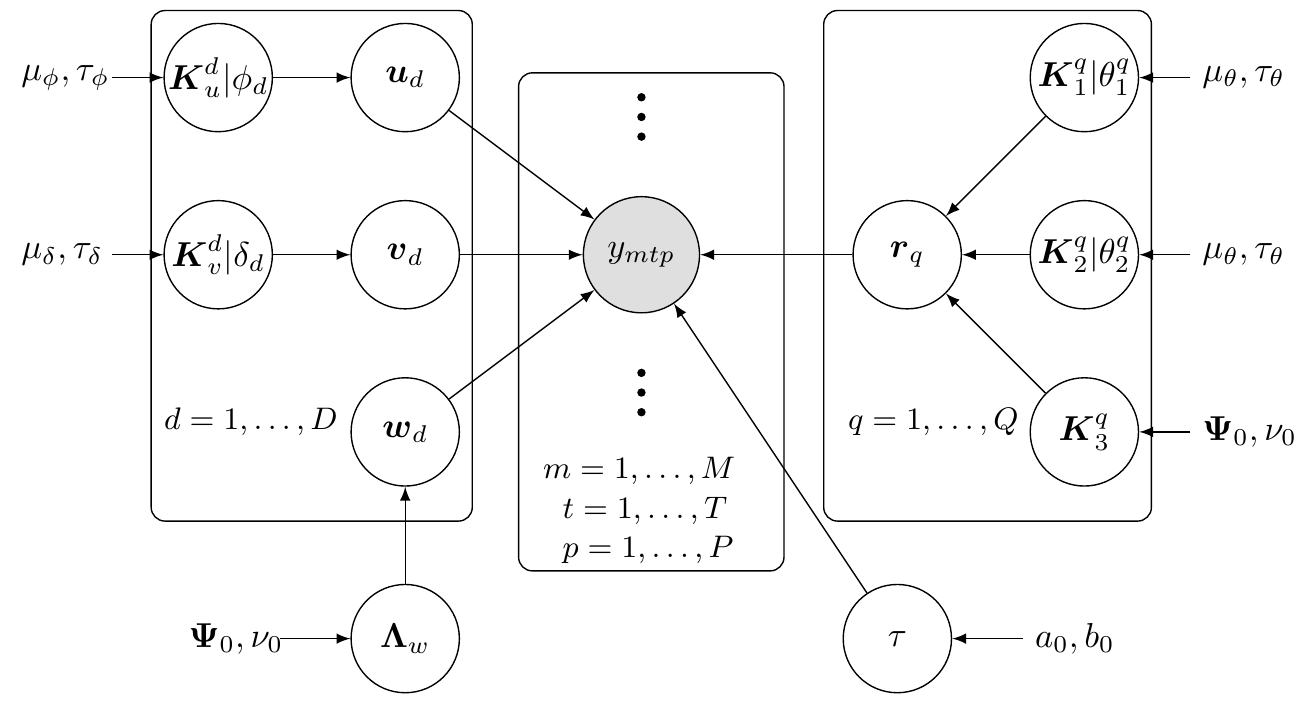}
\caption{Graphical model of BCKL.}
\label{fig_graphicalBCKL}
\end{figure}

In this section, we build a complementary model for multidimensional data by combining Bayesian kernelized tensor factorization with local spatiotemporal GP regression, in which both the global patterns and the local structures of the data can be effectively characterized. We refer to the proposed model as {\textit{Bayesian Complementary Kernelized Learning}} (BCKL). In the following of this section, we describe BCKL on a third-order tensor structure as an example, which can be reduced to matrices and straightforwardly extended to higher-order tensors. We denote by $\boldsymbol{\mathcal{Y}}\in\mathbb{R}^{M\times T\times P}$ an incomplete third-order tensor (\textit{space }$\times$\textit{ time }$\times$\textit{ variable}): the entries in $\boldsymbol{\mathcal{Y}}$ are defined on a 3D space $S_1\times S_2\times S_3=\{\left(s_1,s_2,s_3\right):s_1\in\mathcal{S}_1,s_2\in\mathcal{S}_2,s_3\in\mathcal{S}_3\}$, $\mathcal{S}_1,\mathcal{S}_2,\mathcal{S}_3$ contain $M$, $T$ ,$P$ coordinates, respectively, and $y_{mtp}$ is observed if $(m,t,p)\in\Omega$. We assume $\boldsymbol{\mathcal{Y}}$ is constructed by three component tensors of the same size as $\boldsymbol{\mathcal{Y}}$: 
\begin{equation} \label{Eq_main}
\boldsymbol{\mathcal{Y}}=\boldsymbol{\mathcal{X}}+\boldsymbol{\mathcal{R}}+\boldsymbol{\mathcal{E}},
\end{equation}
where $\boldsymbol{\mathcal{X}}$ represents the latent global tensor capturing long-range/global correlations of the data, $\boldsymbol{\mathcal{R}}$ denotes the latent local tensor and is generated to describe short-range variations in the data, and $\boldsymbol{\mathcal{E}}$ denotes white noise. The graphical model of the BCKL framework is illustrated in Fig.~\ref{fig_graphicalBCKL}.

We model the latent global variable $\boldsymbol{\mathcal{X}}$ by Bayesian kernelized CANDECOMP/PARAFAC (CP) decomposition:
\begin{equation} \label{Eq_GlobalCP}
\boldsymbol{\mathcal{X}}=\sum_{d=1}^{D}\boldsymbol{u}_{d}\circ\boldsymbol{v}_{d}\circ\boldsymbol{w}_{d},
\end{equation}
where $D$ is the tensor CP rank, $\boldsymbol{u}_{d}\in\mathbb{R}^{M}$, $\boldsymbol{v}_{d}\in\mathbb{R}^{T}$, and $\boldsymbol{w}_{d}\in\mathbb{R}^{P}$ are the $d$th column of the decomposed factor matrices $\boldsymbol{U}\in\mathbb{R}^{M\times D}$, $\boldsymbol{V}\in\mathbb{R}^{T\times D}$, and $\boldsymbol{W}\in\mathbb{R}^{P\times D}$, respectively. Each column of the factors $\boldsymbol{U}$ and $\boldsymbol{V}$ is assumed to follow a GP prior:
\begin{equation}
\begin{aligned}
\boldsymbol{u}_{d}&\sim\mathcal{N}\left(\boldsymbol{0},\boldsymbol{K}_{u}^{d}\right),~d=1,2,\ldots,D, \\
\boldsymbol{v}_{d}&\sim\mathcal{N}\left(\boldsymbol{0},\boldsymbol{K}_{v}^{d}\right),~d=1,2,\ldots,D, \\
\end{aligned}
\end{equation}
where $\boldsymbol{K}_{u}^{d}\in\mathbb{R}^{M\times M}$ and $\boldsymbol{K}_{v}^{d}\in\mathbb{R}^{T\times T}$ are covariance matrices obtained from valid kernel functions. Since the factorization model in Eq.~\eqref{Eq_GlobalCP} identifies $\boldsymbol{u}_{d}\circ\boldsymbol{v}_{d}\circ\boldsymbol{w}_{d}$ only, we fix the variances of $\boldsymbol{K}_{u}^{d}$ and $\boldsymbol{K}_{v}^{d}$ to one, and capture the variance/magnitude of $\boldsymbol{\mathcal{X}}$ through $\boldsymbol{w}_{d}$. The kernel functions of $\boldsymbol{K}_{u}^{d}$ and $\boldsymbol{K}_{v}^{d}$, for which we only need to learn the length-scale hyperparameters, are defined by $k_u\left(\Delta_1;\phi_d\right)$ and $k_v\left(\Delta_2;\delta_d\right)$, respectively, with $\Delta_1$ and $\Delta_2$ denoting the distances in space and time, respectively, and $\phi_d$ and $\delta_d$ being the kernel length-scale hyperparameters. To ensure the positivity of the kernel hyperparameters, we use their log-transformed values during inference, and place normal hyper-priors on the transformed variables, i.e., $\log\left(\phi_d\right)\sim\mathcal{N}\left(\mu_{\phi},\tau_{\phi}^{-1}\right)$ and $\log\left(\delta_d\right)\sim\mathcal{N}\left(\mu_{\delta},\tau_{\delta}^{-1}\right)$. For the factor matrix $\boldsymbol{W}$, we assume an identical multivariate normal prior to each column:
\begin{equation}
\begin{aligned}
\boldsymbol{w}_{d}\sim\mathcal{N}\left(\boldsymbol{0},\boldsymbol
{\Lambda}_{w}^{-1}\right),~d=1,2,\ldots,D,
\end{aligned}
\end{equation}
and place a conjugate Wishart prior on the precision matrix, i.e., $\boldsymbol{\Lambda}_{w}\sim\mathcal{W}\left(\boldsymbol{\Psi}_0,\nu_0\right)$, where $\boldsymbol{\Psi}_0\in\mathbb{R}^{P\times P}$ is the scale matrix and $\nu_0$ defines the degrees of freedom. This specification can be considered a multidimensional/spatiotemporal extension of the kernelized matrix factorization \cite{gamerman2008spatial,luttinen2009variational,lei2022bayesian}.

The local component $\boldsymbol{\mathcal{R}}$ is defined on the same grid space as $\boldsymbol{\mathcal{Y}}$. Although kernels in the product form are suitable for high-dimensional datasets, the assumption of separability among different input dimensions is too restrictive for real-world datasets \cite{stein2005space, fuentes2006testing}. To overcome the shortcomings of separable kernels, we construct a non-separable kernel for $\boldsymbol{\mathcal{R}}$ by summing $Q$ product kernel functions \cite{de2001space}, resulting in a sum of separable kernel functions. Usually, a small $Q$, e.g., $Q=2$, could be sufficient to capture the nonseparability of the data \cite{luttinen2012efficient}. Assuming that $\boldsymbol{\mathcal{Y}}$ represents a multivariate spatiotemporal process, the kernel for $\boldsymbol{\mathcal{R}}$ between two inputs $\boldsymbol{s}=\left(s_1,s_2,s_3\right)$ and $\boldsymbol{s}'=\left(s_1',s_2',s_3'\right)$ can be written as: 
\begin{equation}
\begin{aligned}
k_{\mathcal{R}}&\left(\left(s_1,s_2,s_3\right),\left(s_1',s_2',s_3'\right);\boldsymbol{\theta}_{\mathcal{R}}\right)=\\
&\sum_{q=1}^{Q}k_1^{q}\left(\Delta_1;\theta_1^q\right)k_2^{q}\left(\Delta_2;\theta_2^q\right)\boldsymbol{K}_3^{q}\left(s_3,s_3'\right),
\end{aligned}
\end{equation}
where $k_1^q$ and $k_2^q$ are two kernel functions defined for the spatial and temporal dimensions, respectively, with $\Delta_1,\Delta_2$ being the distance between the inputs and $\theta_1^q,\theta_2^q$ being the kernel hyperparameters, and $\boldsymbol{K}_3^q$ is a $P\times P$ symmetric positive-definite matrix capturing the relationships among variables. Hyperparameters of this kernel are $\boldsymbol{\theta}_{\mathcal{R}}=\{\theta_1^q,\theta_2^q,\boldsymbol{K}_3^q\}_{q=1}^Q$. As the magnitudes of the three components interact with each other in a similar way as in $\boldsymbol{u}_{d}$, $\boldsymbol{v}_{d}$ and $\boldsymbol{w}_{d}$, we assume $k_1^q$ and $k_2^q$ are kernels with variances being one with only length-scales $\theta_1^q$ and $\theta_2^q$ to be learned, and use $\boldsymbol{K}_3^q$ to capture the magnitude of $\boldsymbol{\mathcal{R}}$. Given the assumption of short-scale correlations, we construct $k_1^{q}$ and $k_2^q$ using compactly supported tapered covariance functions and place an inverse Wishart prior for variable covariance  $\boldsymbol{K}_{3}^{q}\sim\mathcal{W}^{-1}\left(\boldsymbol{\Psi}_{0}^{-1},\nu_0\right)$. In applications where correlations among variables can be ignored, we can simplify $\boldsymbol{K}_3^q$ to a diagonal matrix, which offers additional computational gains.

Based on the above specifications, the covariance matrix of $\operatorname{vec}\left(\boldsymbol{\mathcal{R}}\right)$ becomes:
\begin{equation}
\boldsymbol{K}_{\mathcal{R}}=\sum_{q=1}^{Q}\boldsymbol{K}_{r}^{q},\ \text{  } \ 
\boldsymbol{K}_{r}^{q}=\boldsymbol{K}_{3}^{q}\otimes\boldsymbol{K}_{2}^{q}\otimes\boldsymbol{K}_{1}^{q},
\end{equation}
where $\boldsymbol{K}_{1}^{q}\in\mathbb{R}^{M\times M}$ and $\boldsymbol{K}_{2}^{q}\in\mathbb{R}^{T\times T}$ are computed from kernel $k_{1}^{q}$ and $k_{2}^{q}$, respectively. The prior of the vectorized $\boldsymbol{\mathcal{R}}$, i.e., $\operatorname{vec}(\boldsymbol{\mathcal{R}})$, can be written as:
\begin{equation}
\operatorname{vec}(\boldsymbol{\mathcal{R}})\sim\mathcal{N}\left(\boldsymbol{0},\boldsymbol{K}_{\mathcal{R}}\right),
\end{equation}
where $\operatorname{vec}(\boldsymbol{\mathcal{R}})=\sum\nolimits_{q=1}^{Q}\boldsymbol{r}_{q}$ and $\boldsymbol{r}_{q}\sim\mathcal{N}\left(\boldsymbol{0},\boldsymbol{K}_{r}^{q}\right)$. For the hyper-prior of kernel hyperparameters $\{\theta_1^q,\theta_2^q\}$ in $\boldsymbol{\theta}_{\mathcal{R}}$, we assume the log-transformed variable follows a normal prior, i.e., $\log{(\theta)}\sim\mathcal{N}\left(\mu_{\theta},\tau_{\theta}^{-1}\right)$, in a similar way to learning $\phi_d$ and $\delta_d$.

The last noise term $\boldsymbol{\mathcal{E}}$ is assumed to be an i.i.d. white noise process for observed entries, i.e., $\mathcal{N}\left(0,\tau^{-1}\right)$. We place a conjugate Gamma prior on the precision parameter $\tau$, that is $\tau\sim\text{Gamma}(a_0,b_0)$. Each observed entry in $\boldsymbol{\mathcal{Y}}$ has the following distribution:
\begin{equation}
y_{mtp}\sim\mathcal{N}\left(\sum\limits_{d=1}^{D}u_{md}v_{td}w_{pd}+r_{mtp},\tau^{-1}\right),~\forall(m,t,p)\in\Omega,
\end{equation}
and equivalently, we have
\begin{equation} \label{Eq_full_kernel}
\left.\operatorname{vec}(\boldsymbol{\mathcal{Y}})_{\Omega} \right|\boldsymbol{U},\boldsymbol{V},\boldsymbol{W},\tau,\boldsymbol{\theta}_{\mathcal{R}}\sim \mathcal{N}\left(\operatorname{vec}(\boldsymbol{\mathcal{X}})_{\Omega},\left[\boldsymbol{K}_{\mathcal{R}}\right]_{\Omega}+\tau^{-1}\boldsymbol{I}_{|\Omega|}\right),
\end{equation}
where $\boldsymbol{\mathcal{X}}=\sum_{d=1}^{D} \boldsymbol{u}_{d}\circ\boldsymbol{v}_{d}\circ\boldsymbol{w}_{d}$, and the operator $(\cdot)_{\Omega}$ denotes the projection of a full vector or a full covariance matrix onto the observed indices. The global term $\boldsymbol{\mathcal{X}}$ modeled by tensor factorization becomes a flexible and effective mean/trend term that can capture higher-order interactions among different dimensions, ensuring that the residual process can be better characterized by a local short-scale GP. With the use of GP priors on the columns of $\boldsymbol{U}$ and $\boldsymbol{V}$, we can also perform interpolation such as in kernelized factorization models \cite{gamerman2008spatial,luttinen2009variational,lei2022bayesian}. In addition, as a reduced rank model, tensor factorization provides a flexible approach to model nonstationary and nonseparable processes. Note that although we assume i.i.d. noise here, it is easy to introduce space-specific, time-specific, or variable-specific error distributions as long as there is sufficient data to model the noise processes. For example, we can learn different $\tau_p$ for $p=1,\ldots,P$ using data from the $p$th frontal slice of the whole data tensor.

\subsection{Model inference} \label{subsec:mcmc}

In this subsection, we introduce an efficient Gibbs sampling algorithm for model inference.

\subsubsection{Sampling factors of the global tensor}

Since the prior distributions and the likelihoods of the factors decomposed from the global tensor, i.e.,  $\left\{\boldsymbol{u}_{d},\boldsymbol{v}_{d},\boldsymbol{w}_{d}\right\}$, are both assumed to be Gaussian, the posterior of each factor matrix still follows a Gaussian distribution. Let $\boldsymbol{\mathcal{Y}}_{\backslash\mathcal{R}}^{d}=\boldsymbol{\mathcal{Y}}-\boldsymbol{\mathcal{R}}-\sum_{h:h=[1,D]\backslash d}\boldsymbol{u}_{h}\circ\boldsymbol{v}_{h}\circ\boldsymbol{w}_{h}$ for $d=1,\cdots,D$, and $\boldsymbol{\mathcal{O}}\in\mathbb{R}^{M\times T\times P}$ be a binary indicator tensor with $o_{mtp}=1$ if $(m,t,p)\in\Omega$ and $o_{mtp}=0$ otherwise. The mode-1 unfolding of $\boldsymbol{\mathcal{Y}}_{\backslash\mathcal{R}}^{d}$, denoted by $\boldsymbol{Y}_{\backslash\mathcal{R}(1)}^{d}\in\mathbb{R}^{M\times(TP)}$, can be represented as $\boldsymbol{Y}_{\backslash\mathcal{R}(1)}^{d}=\boldsymbol{u}_{d}(\boldsymbol{w}_{d}\otimes\boldsymbol{v}_{d})^{\top}+\boldsymbol{E}_{(1)}$, where $\boldsymbol{E}_{(1)}$ is the mode-1 unfolding of the noise tensor $\boldsymbol{\mathcal{E}}$. Based on the identity $\operatorname{vec}\left(\boldsymbol{A}\boldsymbol{X}\boldsymbol{B}\right)=\left(\boldsymbol{B}^{\top}\otimes\boldsymbol{A}\right)\operatorname{vec}(\boldsymbol{X})$, we have $\operatorname{vec}\left(\boldsymbol{Y}_{\backslash\mathcal{R}(1)}^{d}\right)_{\Omega}=\left[\left((\boldsymbol{w}_{d}\otimes\boldsymbol{v}_{d})\otimes\boldsymbol{I}_{M}\right)\boldsymbol{u}_{d}+\operatorname{vec}\left(\boldsymbol{E}_{(1)}\right)\right]_{\Omega}$. With the Gaussian prior for each column, the conditional distribution of $\boldsymbol{u}_{d}\in\mathbb{R}^{M}:d=1,\ldots,D$, is also a Gaussian as below:
\begin{equation}
\begin{aligned}
p\left(\boldsymbol{u}_{d}\given-\right)&=\mathcal{N}\left(\boldsymbol{u}_{d}\given\boldsymbol{\mu}_{ud}^{*},(\boldsymbol{\Lambda}_{ud}^{*})^{-1}\right), \\
\boldsymbol{\mu}_{ud}^{*}&=\tau(\boldsymbol{\Lambda}_{ud}^{*})^{-1}(\boldsymbol{Y}_{\backslash\mathcal{R}(1)}^{d}\circledast\boldsymbol{O}_{(1)})(\boldsymbol{w}_{d}\otimes\boldsymbol{v}_{d}), \\
\boldsymbol{\Lambda}_{ud}^{*}&=\tau\boldsymbol{H}^{\top}\boldsymbol{H}+\left(\boldsymbol{K}_{u}^{d}\right)^{-1}, \\
\boldsymbol{H}&=\boldsymbol{O}_{1}\left((\boldsymbol{w}_{d}\otimes\boldsymbol{v}_{d})\otimes\boldsymbol{I}_{M}\right),
\end{aligned}
\end{equation}
where $\boldsymbol{O}_{(1)}$ is the mode-1 unfolding of $\boldsymbol{\mathcal{O}}$, and $\boldsymbol{O}_{1}\in\mathbb{R}^{|\Omega|\times (MTP)}$ is a binary matrix formed by removing the rows corresponding to zero values in $\operatorname{vec}\left(\boldsymbol{O}_{(1)}\right)$ from a $MTP\times MTP$ identity matrix. Note that $\boldsymbol{H}^{\top}\boldsymbol{H}=\text{diag}\left(\boldsymbol{O}_{(1)}(\boldsymbol{w}_{d}\otimes\boldsymbol{v}_{d})^2\right)$ is a diagonal matrix and can be efficiently computed. When $M$ is large, further computational gains can be obtained by scalable GP models, such as using sparse approximation (predictive process) or assuming a sparse precision matrix for the latent factors so that $\left(\boldsymbol{K}_{u}^{d}\right)^{-1}$ is available by default.

Similarly, we get the conditional distributions of $\boldsymbol{v}_{d}$ and $\boldsymbol{w}_{d}$ for $d=1,\ldots,D$:
\begin{equation}
\begin{aligned}
p\left(\boldsymbol{v}_{d}\given-\right)&=\mathcal{N}\left(\boldsymbol{v}_{d}\given\boldsymbol{\mu}_{vd}^{*},(\boldsymbol{\Lambda}_{vd}^{*})^{-1}\right), \\
\boldsymbol{\mu}_{vd}^{*}&=\tau(\boldsymbol{\Lambda}_{vd}^{*})^{-1}(\boldsymbol{Y}_{\backslash\mathcal{R}(2)}^{d}\circledast\boldsymbol{O}_{(2)})(\boldsymbol{w}_{d}\otimes\boldsymbol{u}_{d}), \\
\boldsymbol{\Lambda}_{vd}^{*}&=\tau\boldsymbol{H}^{\top}\boldsymbol{H}+\left(\boldsymbol{K}_{v}^{d}\right)^{-1}, \\
\boldsymbol{H}&=\boldsymbol{O}_{2}\left((\boldsymbol{w}_{d}\otimes\boldsymbol{u}_{d})\otimes\boldsymbol{I}_{T}\right), \\
\end{aligned}
\end{equation}
and
\begin{equation}
\begin{aligned}
p\left(\boldsymbol{w}_{d}\given-\right)&=\mathcal{N}\left(\boldsymbol{w}_{d}\given\boldsymbol{\mu}_{wd}^{*},(\boldsymbol{\Lambda}_{wd}^{*})^{-1}\right), \\
\boldsymbol{\mu}_{wd}^{*}&=\tau(\boldsymbol{\Lambda}_{wd}^{*})^{-1}(\boldsymbol{Y}_{\backslash\mathcal{R}(3)}^{d}\circledast\boldsymbol{O}_{(3)})(\boldsymbol{v}_{d}\otimes\boldsymbol{u}_{d}), \\
\boldsymbol{\Lambda}_{wd}^{*}&=\tau\boldsymbol{H}^{\top}\boldsymbol{H}+\boldsymbol{\Lambda}_{w}, \\
\boldsymbol{H}&=\boldsymbol{O}_{3}\left((\boldsymbol{v}_{d}\otimes\boldsymbol{u}_{d})\otimes\boldsymbol{I}_{P}\right),
\end{aligned}
\end{equation}
where $\boldsymbol{Y}_{\backslash\mathcal{R}(2)}^{d},\boldsymbol{Y}_{\backslash\mathcal{R}(3)}^{d}$, and $\boldsymbol{O}_{(2)},\boldsymbol{O}_{(3)}$ are the mode-2 and mode-3 unfoldings of $\boldsymbol{\mathcal{Y}}_{\backslash\mathcal{R}}^{d}$ and $\boldsymbol{\mathcal{O}}$, respectively; $\boldsymbol{O}_2$ and $\boldsymbol{O}_3$ are binary matrices of size $|\Omega|\times MTP$ obtained by removing the rows corresponding to zeros in $\operatorname{vec}\left(\boldsymbol{O}_{(2)}\right)$ and $\operatorname{vec}\left(\boldsymbol{O}_{(3)}\right)$ from $\boldsymbol{I}_{MTP}$, respectively. After sampling $\boldsymbol{U}$, $\boldsymbol{V}$, and $\boldsymbol{W}$, the global component tensor $\boldsymbol{\mathcal{X}}$ can be calculated using Eq.~\eqref{Eq_GlobalCP}.

\subsubsection{Sampling hyperparameters of the global component}

The hyperparameters of the global component $\boldsymbol{\mathcal{X}}$ are defined as $\boldsymbol{\Theta}_{\mathcal{X}}=\{\phi_{d},\delta_d:d=[1,D],\boldsymbol{\Lambda}_{w}\}$, which include the kernel hyperparameters of $\boldsymbol{U}$ and $\boldsymbol{V}$, i.e., $\{\phi_{d},\delta_d\}$, and the precision matrix of $\boldsymbol{W}$, i.e., $\boldsymbol{\Lambda}_{w}$. We learn the kernel hyperparameters $\{\phi_{d}, \delta_d\}$ from their marginal posteriors based on slice sampling, and update $\boldsymbol{\Lambda}_{w}$ using the Gibbs sampling. Specifically, the marginal likelihood of $\phi_d$ is:
\begin{equation} \label{Eq_TgMarginal}
\begin{aligned}
&\log {p\left(\left(\boldsymbol{y}_{\backslash\mathcal{R}}^{d}\right)_{\Omega}\given\phi_{d}\right)}\propto\\
&-\frac{1}{2}\left(\boldsymbol{y}_{\backslash\mathcal{R}}^{d}\right)_{\Omega}^{\top}\boldsymbol{\Sigma}_{\left.(\boldsymbol{y}_{\backslash\mathcal{R}}^{d})_{\Omega}\right|\phi_{d}}^{-1}\left(\boldsymbol{y}_{\backslash\mathcal{R}}^{d}\right)_{\Omega}-\frac{1}{2}\log{\left|\boldsymbol{\Sigma}_{\left.(\boldsymbol{y}_{\backslash\mathcal{R}}^{d})_{\Omega}\right|\phi_{d}}\right|}, \\
\end{aligned}
\end{equation}
where $\left(\boldsymbol{y}_{\backslash\mathcal{R}}^{d}\right)_{\Omega}=\boldsymbol{O}_{1}\operatorname{vec}\left(\boldsymbol{Y}_{\backslash\mathcal{R}(1)}^{d}\right)$, and $\boldsymbol{\Sigma}_{\left.(\boldsymbol{y}_{\backslash\mathcal{R}}^{d})_{\Omega}\right|\phi_{d}}=\boldsymbol{H}\boldsymbol{K}_{u}^{d}\boldsymbol{H}^{\top}+\tau^{-1}\boldsymbol{I}_{|\Omega|}$ with $\boldsymbol{H}=\boldsymbol{O}_{1}\left((\boldsymbol{w}_{d}\otimes\boldsymbol{v}_{d})\otimes\boldsymbol{I}_{M}\right)$.

In computing Eq.~\eqref{Eq_TgMarginal}, we can leverage Sherman-Woodbury-Morrison-type computations \cite{harville1998matrix, banerjee2014hierarchical}, since we have $M\ll |\Omega|$ in general. Particularly, the inverse can be evaluated as
\begin{equation}
\begin{aligned}
\boldsymbol{\Sigma}_{\left.(\boldsymbol{y}_{\backslash\mathcal{R}}^{d})_{\Omega}\right|\phi_{d}}^{-1} &= \left(\tau^{-1}\boldsymbol{I}_{|\Omega|}+\boldsymbol{H}\boldsymbol{K}_{u}^{d}\boldsymbol{H}^{\top}\right)^{-1}\\
&=\tau\boldsymbol{I}_{|\Omega|}-\tau^2\boldsymbol{H}\left(\left(\boldsymbol{K}_{u}^{d}\right)^{-1}+\tau\boldsymbol{H}^{\top}\boldsymbol{H}\right)^{-1}\boldsymbol{H}^{\top},
\end{aligned}
\end{equation}
and the determinant can be computed with
\begin{equation}
\begin{aligned}
\left|\boldsymbol{\Sigma}_{\left.(\boldsymbol{y}_{\backslash\mathcal{R}}^{d})_{\Omega}\right|\phi_{d}}\right|&=\left|\tau^{-1}\boldsymbol{I}_{|\Omega|}+\boldsymbol{H}\boldsymbol{K}_{u}^{d}\boldsymbol{H}^{\top}\right|\\
&=\left|\left(\boldsymbol{K}_{u}^{d}\right)^{-1}+\tau\boldsymbol{H}^{\top}\boldsymbol{H}\right|\cdot\left|\boldsymbol{K}_{u}^{d}\right|\cdot\left|\tau^{-1}\boldsymbol{I}_{|\Omega|}\right| \\
&\propto\left|\left(\boldsymbol{K}_{u}^{d}\right)^{-1}+\tau\boldsymbol{H}^{\top}\boldsymbol{H}\right|\cdot\left|\boldsymbol{K}_{u}^{d}\right|.
\end{aligned}
\end{equation}
This avoids solving the inverse and determinant of the large $|\Omega|\times|\Omega|$ covariance matrix $\boldsymbol{\Sigma}_{\left.(\boldsymbol{y}_{\backslash\mathcal{R}}^{d})_{\Omega}\right|\phi_{d}}$, and reduces the computational cost from $\mathcal{O}(|\Omega|^{3})$ to $\mathcal{O}(M^3)$. The log posterior of $\log{p\left(\phi_d\given-\right)}$ equals the sum of  $\log{p\left(\left(\boldsymbol{y}_{\backslash\mathcal{R}}^{d}\right)_{\Omega}\given\phi_{d}\right)}$ and $\log{p\left(\phi_d\right)}$ up to a constant. The marginal posterior of $\delta_d$ can be computed similarly. We use the robust slice sampling method to sample $\phi_d$ and $\delta_d$ from their posterior distributions \cite{murray2010slice}. The implementation of slice sampling is summarized in Algorithm~\ref{alg_RHyperSample}.

As for the hyperparameter of $\boldsymbol{W}$, i.e., $\boldsymbol{\Lambda}_{w}$, the posterior distribution is given by a Wishart distribution $p\left(\boldsymbol{\Lambda}_{w}\given-\right)=\mathcal{W}\left(\boldsymbol{\Lambda}_{w}\given\boldsymbol{\Psi}^{*},\nu^{*}\right)$, where $\left(\boldsymbol{\Psi}^{*}\right)^{-1}=\boldsymbol{W}\boldsymbol{W}^{\top}+\boldsymbol{\Psi}_{0}^{-1}$ and $\nu^{*}=\nu_0+D$.

\subsubsection{Sampling the local component}

Following the theory of GP regression, the posteriors of the local components $\left\{\boldsymbol{r}_{q}\in\mathbb{R}^{MTP}:q\in[1,Q]\right\}$ are Gaussian distributions. For $\boldsymbol{r}_{q}$, we denote its mean and covariance by $\boldsymbol{\mu}_{\left.\boldsymbol{r}_{q}\right|-}$ and $\boldsymbol{\Sigma}_{\left.\boldsymbol{r}_{q}\right|-}$, respectively. Instead of directly sampling $\boldsymbol{r}_{q}$ from $\mathcal{N}\left(\boldsymbol{r}_{q}\given\boldsymbol{\mu}_{\left.\boldsymbol{r}_{q}\right|-},\boldsymbol{\Sigma}_{\left.\boldsymbol{r}_{q}\right|-}\right)$ which involves the calculation of an ultra high-dimensional covariance matrix, we draw samples of $\boldsymbol{r}_{q}$ in a more efficient way as shown in Lemma 1 following \cite{luttinen2012efficient}.

\textit{
Lemma 1: Let $\boldsymbol{x}$ and $\boldsymbol{y}$ be multivariate Gaussian random vectors:
\begin{equation} \label{EqJoint} \notag
\begin{aligned}
\begin{bmatrix}
\boldsymbol{x} \\
\boldsymbol{y}
\end{bmatrix}\sim\mathcal{N}\left(\begin{bmatrix}
\boldsymbol{\mu}_{x} \\ \boldsymbol{\mu}_{y}
\end{bmatrix},\begin{bmatrix}
\boldsymbol{\Sigma}_{x}~~\boldsymbol{\Sigma}_{\left(x,y\right)} \\
\boldsymbol{\Sigma}_{\left(y,x\right)}~~\boldsymbol{\Sigma}_{y}
\end{bmatrix}\right).
\end{aligned}
\end{equation}
If we can generate samples $\{\Tilde{\boldsymbol{x}},\Tilde{\boldsymbol{y}}\}$ from the joint distribution $p\left(\boldsymbol{x},\boldsymbol{y}\right)$, then the sample $\boldsymbol{x}^{\ast}=\Tilde{\boldsymbol{x}}-\boldsymbol{\Sigma}_{(x,y)}\boldsymbol{\Sigma}_{y}^{-1}\left(\Tilde{\boldsymbol{y}}-\boldsymbol{y}\right)$ follows the conditional distribution $p\left(\boldsymbol{x}\given\boldsymbol{y}\right)$, where $\boldsymbol{\Sigma}_{(x,y)}$ and $\boldsymbol{\Sigma}_{y}$ are the covariance matrix between $\boldsymbol{x}$ and $\boldsymbol{y}$, and the covariance of $\boldsymbol{y}$, respectively.
}

Let $\boldsymbol{\mathcal{Y}}_{\backslash\mathcal{X}}=\boldsymbol{\mathcal{Y}}-\boldsymbol{\mathcal{X}}$, $\boldsymbol{Y}_{\backslash\mathcal{X}(1)}$ be the mode-1 unfolding of $\boldsymbol{\mathcal{Y}}_{\backslash\mathcal{X}}$, and $\boldsymbol{y}_{\backslash\mathcal{X}}=\operatorname{vec}\left(\boldsymbol{Y}_{\backslash\mathcal{X}(1)}\right)$. 
Given that 
\begin{equation} \notag
\begin{aligned}
&\boldsymbol{r}_{q}\sim\mathcal{N}\left(\boldsymbol{0},\boldsymbol{K}_{r}^{q}\right),~q=1,\ldots,Q, \\    &\left(\boldsymbol{y}_{\backslash\mathcal{X}}\right)_{\Omega}\sim\mathcal{N}\left(\boldsymbol{O}_{1}\sum_{q=1}^{Q}\boldsymbol{r}_{q},\tau^{-1}\boldsymbol{I}_{|\Omega|}\right),
\end{aligned}
\end{equation}
where $\left(\boldsymbol{y}_{\backslash\mathcal{X}}\right)_{\Omega}=\boldsymbol{O}_{1}\boldsymbol{y}_{\backslash\mathcal{X}}$. Thus, the joint distribution of $\boldsymbol{r}_{q}$ and $\left(\boldsymbol{y}_{\backslash\mathcal{X}}\right)_{\Omega}$ is a multivariate Gaussian distribution: if we have samples drawn from this joint distribution, for example $\left\{\Tilde{\boldsymbol{r}}_{q},\left(\Tilde{\boldsymbol{y}}_{\backslash\mathcal{X}}\right)_{\Omega}\right\}$, then the variable $\boldsymbol{r}_{q}^{\ast}=\Tilde{\boldsymbol{r}}_{q}-\boldsymbol{\Sigma}_{\left(r_{q},(y_{\backslash\mathcal{X}})_{\Omega}\right)}\boldsymbol{\Sigma}_{(y_{\backslash\mathcal{X}})_{\Omega}}^{-1}\left(\left(\Tilde{\boldsymbol{y}}_{\backslash\mathcal{X}}\right)_{\Omega}-\left(\boldsymbol{y}_{\backslash\mathcal{X}}\right)_{\Omega}\right)$ is a sample from the conditional distribution $p\left(\boldsymbol{r}_{q}\given\left(\boldsymbol{y}_{\backslash\mathcal{X}}\right)_{\Omega}\right)$.

We first draw samples $\left\{\Tilde{\boldsymbol{r}}_{q},\Tilde{\boldsymbol{y}}_{\backslash\mathcal{X}}\right\}$ from the joint distribution $p\left(\boldsymbol{r}_{q},\boldsymbol{y}_{\backslash\mathcal{X}}\right)$ through:
\begin{equation} \label{EqJointsample}
\begin{aligned}
\Tilde{\boldsymbol{r}}_{q}&=\operatorname{vec}\left(\boldsymbol{L}_{1}^{q}\boldsymbol{Z}_{q}\left(\boldsymbol{L}_{3}^{q}\otimes\boldsymbol{L}_{2}^{q}\right)^{\top}\right),~q=1,\ldots,Q, \\
\Tilde{\boldsymbol{y}}_{\backslash\mathcal{X}}&=\sum_{q=1}^{Q}\Tilde{\boldsymbol{r}}_{q}+\boldsymbol{z},
\end{aligned}
\end{equation}
where $\boldsymbol{L}_{1}^{q}$, $\boldsymbol{L}_{2}^{q}$, and $\boldsymbol{L}_{3}^{q}$ are the Cholesky factor matrices of $\boldsymbol{K}_{1}^{q}$, $\boldsymbol{K}_{2}^{q}$, and $\boldsymbol{K}_{3}^{q}$, respectively, $\boldsymbol{Z}_{q}\in\mathbb{R}^{M\times (TP)}$ is a matrix sampled from the standard normal distribution, and $\boldsymbol{z}\in\mathbb{R}^{MTP}$ is a vector sampled from $\mathcal{N}\left(\boldsymbol{0},\tau^{-1}\boldsymbol{I}_{MTP}\right)$. The sampling in Eq.~\eqref{EqJointsample} is efficient because $\{\boldsymbol{L}_{1}^{q}\in\mathbb{R}^{M\times M},\boldsymbol{L}_{2}^{q}\in\mathbb{R}^{T\times T},\boldsymbol{L}_{3}^{q}\in\mathbb{R}^{P\times P}\}$ are sparse lower triangular matrices of small sizes. One can obtain the sample $\boldsymbol{r}_{q}$ that follows the conditional distribution $p\left(\boldsymbol{r}_{q}\given\left(\boldsymbol{y}_{\backslash\mathcal{X}}\right)_{\Omega}\right)$, written as
\begin{equation} \label{Eq_rSample}
\begin{aligned}
&\boldsymbol{r}_{q}=\Tilde{\boldsymbol{r}}_{q}-\boldsymbol{K}_{r}^{q}\boldsymbol{O}_{1}^{\top}\times\\
&\underbrace{\left(\boldsymbol{O}_{1}\left(\sum_{q=1}^{Q}\boldsymbol{K}_{r}^{q}\right)\boldsymbol{O}_{1}^{\top}+\tau^{-1}\boldsymbol{I}_{|\Omega|}\right)^{-1}\left(\boldsymbol{O}_{1}\Tilde{\boldsymbol{y}}_{\backslash\mathcal{X}}-\left(\boldsymbol{y}_{\backslash\mathcal{X}}\right)_{\Omega}\right)}_{\boldsymbol{c}},
\end{aligned}
\end{equation}
where the term $\boldsymbol{c}\in\mathbb{R}^{|\Omega|\times 1}$ is the same for all $\boldsymbol{r}_{q}:q\in[1,Q]$, and only needs to be computed once.

Following the approach introduced in \cite{wilson2014fast}, we further introduce $(MTP-|\Omega|)$ imaginary observations $\boldsymbol{y}_{I}\sim\mathcal{N}\left(\boldsymbol{0},\tau_{I}^{-1}\boldsymbol{I}\right)$ for the data at missing positions to reduce the computation complexity, where $\tau_{I}\rightarrow 0$. Denoting $\boldsymbol{y}_{O}=\boldsymbol{O}_{1}\Tilde{\boldsymbol{y}}_{\backslash\mathcal{X}}-\left(\boldsymbol{y}_{\backslash\mathcal{X}}\right)_{\Omega}$ and placing $\boldsymbol{y}_{O}$ on the observed data points. The concatenated data and noise are defined as $\boldsymbol{y}'=\boldsymbol{O}_{1}^{\top}\boldsymbol{y}_{O}+(\boldsymbol{O}_{1}^{c})^{\top}\boldsymbol{y}_{I}$ and $\boldsymbol{E}'=\tau^{-1}\text{diag}(\operatorname{vec}(\boldsymbol{\mathcal{O}}))+\tau_{I}^{-1}\text{diag}(\operatorname{vec}(\boldsymbol{\mathcal{O}}^c))$, respectively, where $\boldsymbol{O}_{1}^{c}\in\mathbb{R}^{(MTP-|\Omega|)\times (MTP)}$ is a binary matrix constructed by removing rows corresponding to nonzero values in $\operatorname{vec}(\boldsymbol{O}_{(1)})$ from $\boldsymbol{I}_{MTP}$. Note that $\boldsymbol{\mathcal{O}}^{c}\in\mathbb{R}^{M\times T\times P}$ is the complement of $\boldsymbol{\mathcal{O}}$. Since the imaginary observations do not affect the inference when $\tau_{I}\rightarrow0$, the term $\boldsymbol{O}_1^{\top}\boldsymbol{c}$ in Eq.~\eqref{Eq_rSample} can be approximated by $\left(\sum_{q=1}^{Q}\boldsymbol{K}_r^{q}+\boldsymbol{E}'\right)^{-1}\boldsymbol{y}'$ with a small $\tau_I$. We can then sample $\boldsymbol{r}_{q}$ by:
\begin{equation} \label{Eq_rUpdate}
\boldsymbol{r}_{q}=\Tilde{\boldsymbol{r}}_{q}-\boldsymbol{K}_{r}^{q}\left(\sum_{q=1}^{Q}\boldsymbol{K}_{r}^{q}+\boldsymbol{E}'\right)^{-1}\boldsymbol{y}'.
\end{equation}
The linear system in Eq.~\eqref{Eq_rUpdate} (i.e., in the form of $\boldsymbol{K}^{-1}\boldsymbol{y}$) can be efficiently solved using an iterative preconditioned conjugate gradient (PCG) method, where the preconditioner matrix is set as $(\boldsymbol{E}')^{-\frac{1}{2}}$. Note that computing $\left(\sum_{q=1}^{Q}\boldsymbol{K}_{r}^{q}+\boldsymbol{E}'\right)\boldsymbol{y}'$ in PCG is very efficient, as $\boldsymbol{E}'$ is a diagonal matrix and $\left(\sum_{q}\boldsymbol{K}_{r}^{q}\right)\boldsymbol{y}'=\left(\sum_{q}\boldsymbol{K}_3^q\otimes\boldsymbol{K}_{2}^{q}\otimes\boldsymbol{K}_{1}^{q}\right)\boldsymbol{y}'$ can be calculated using Kronecker properties. The local component tensor $\boldsymbol{\mathcal{R}}$ can be then computed as the sum of $\left\{\boldsymbol{r}_{q}:q=[1,Q]\right\}$, that is $\operatorname{vec}(\boldsymbol{\mathcal{R}})=\sum_{q=1}^{Q}\boldsymbol{r}_{q}$.

\subsubsection{Sampling hyperparameters of the local processes}

The kernel hyperparameters of the local component, i.e., $\left\{\boldsymbol{\theta}_{\mathcal{R}}^{q}=\left\{\theta_1^q,\theta_2^q,\boldsymbol{K}_3^q\right\}:q=[1,Q]\right\}$, can be updated by sampling from the marginal posteriors $p\left(\boldsymbol{\theta}_{\mathcal{R}}^{q}\given\boldsymbol{y}_{\backslash\mathcal{X}}^{q}\right)\propto p\left(\boldsymbol{y}_{\backslash\mathcal{X}}^{q}\given\boldsymbol{\theta}_{\mathcal{R}}^{q}\right)p(\boldsymbol{\theta}_{\mathcal{R}}^{q})$ analytically, where $\boldsymbol{y}_{\backslash\mathcal{X}}^{q}=\boldsymbol{y}_{\backslash\mathcal{X}}-\sum_{l:l=[1,Q]\backslash q}\boldsymbol{r}_{l}$ for $q=1,\ldots,Q$. However, approximating the unbiased marginal likelihood $p\left(\boldsymbol{y}_{\backslash\mathcal{X}}^{q}\given\boldsymbol{\theta}_{\mathcal{R}}^{q}\right)$ in covariance tapering would require the full inverse of a $|\Omega|\times|\Omega|$ sparse covariance matrix and leads to prohibitive computational costs \cite{kaufman2008covariance}. To alleviate the computational burden, rather than marginalizing out the latent variables $\boldsymbol{r}_{q}$, we learn $\{\theta_1^q,\theta_2^q\}$ using the likelihood conditioned on a whitened $\boldsymbol{r}_{q}$ based on the whitening strategy proposed in \cite{murray2010slice}. We reparameterize the model by introducing auxiliary variables $\left\{\boldsymbol{G}_{q}\in\mathbb{R}^{M\times(TP)}:q=[1,Q]\right\}$ that satisfy $\boldsymbol{r}_{q}=\operatorname{vec}\left(\boldsymbol{L}_{1}^{q}\boldsymbol{G}_{q}\left(\boldsymbol{L}_{3}^{q}\otimes\boldsymbol{L}_{2}^{q}\right)^{\top}\right)$, and the conditional posterior distribution of $\boldsymbol{\theta}_{\mathcal{R}}^{q}$ becomes:
\begin{equation} \label{Eq_RHyper}
p\left(\boldsymbol{\theta}_{\mathcal{R}}^{q}\given\boldsymbol{G}_{q},\left(\boldsymbol{y}_{\backslash\mathcal{X}}^{q}\right)_{\Omega}\right)\propto p\left(\left(\boldsymbol{y}_{\backslash\mathcal{X}}^{q}\right)_{\Omega}\given\boldsymbol{r}_{q}\left(\boldsymbol{G}_{q},\boldsymbol{\theta}_{\mathcal{R}}^{q}\right)\right)p(\boldsymbol{\theta}_{\mathcal{R}}^{q}).
\end{equation}
Sampling from Eq.~\eqref{Eq_RHyper} is efficient since ${p\left(\left(\boldsymbol{y}_{\backslash\mathcal{X}}^{q}\right)_{\Omega}\given\boldsymbol{r}_{q}\left(\boldsymbol{G}_{q},\boldsymbol{\theta}_{\mathcal{R}}^{q}\right)\right)}$ has a diagonal covariance matrix $\tau^{-1}\boldsymbol{I}_{|\Omega|}$. We use slice sampling operators to update length-scales in $\boldsymbol{\theta}_{\mathcal{R}}^{q}$. The detailed sampling algorithm for $\theta_1^q$ is summarized in Algorithm~\ref{alg_RHyperSample}, and $\theta_2^q$ can be sampled similarly. Note that one can also use a rectangle slice sampling \cite{neal2003slice} to jointly update $\{\theta_1^q,\theta_2^q\}$.

For $\boldsymbol{K}_{3}^{q}$, its posterior conditioned on $\boldsymbol{r}_{q}$ is an inverse Wishart distribution $p\left(\boldsymbol{K}_{3}^{q}\given-\right)=\mathcal{W}^{-1}\left(\boldsymbol{K}_{3}^{q}\given\boldsymbol{\Psi}_{R}^{\ast},\nu_R^{\ast}\right)$, and we update $\boldsymbol{K}_{3}^{q}$ using entries in $\boldsymbol{r}_{q}$ that correspond to the observed data points. Let $\boldsymbol{\omega}$ be a $N_{\omega}\times (MT)$ binary matrix formed by removing rows corresponding to rows of all zeros in $\boldsymbol{O}_{(3)}^{\top}$ from $\boldsymbol{I}_{MT}$, then we have $\boldsymbol{\Psi}_{R}^{\ast}=[\boldsymbol{R}_{q(3)}]_{\Omega}\left[\boldsymbol{K}_2^{q}\otimes\boldsymbol{K}_{1}^{q}\right]_{\Omega}^{-1}[\boldsymbol{R}_{q(3)}]_{\Omega}^{\top}+\boldsymbol{\Psi}_{0}^{-1}$ and $\nu_{R}^{\ast}=\nu_{0}+N_{\omega}$, where $\left[\boldsymbol{K}_2^{q}\otimes\boldsymbol{K}_{1}^{q}\right]_{\Omega}=\boldsymbol{\omega}(\boldsymbol{K}_2^{q}\otimes\boldsymbol{K}_{1}^{q})\boldsymbol{\omega}^{\top}$, $\boldsymbol{R}_{q(3)}^{\top}=\text{reshape}(\boldsymbol{r}_{q},[MT,P])$, $N_{\omega}$ is the number of rows in $\boldsymbol{\omega}$ and $[\boldsymbol{R}_{q(3)}]_{\Omega}^{\top}=\boldsymbol{\omega}\boldsymbol{R}_{q(3)}^{\top}$. It should be noted that the computation of $\boldsymbol{\Psi}_{R}^{\ast}$ involves the inverse of a sparse matrix $\left[\boldsymbol{K}_2^{q}\otimes\boldsymbol{K}_{1}^{q}\right]_{\Omega}$. This calculation can be easily solved using a Cholesky decomposition when the number of missing points is large. On the other hand, when $N_{\omega}$ is large, this procedure can also be accelerated by introducing imaginary observations and utilizing the Kronecker product structure in a similar way as in Eq.~\eqref{Eq_rUpdate}. For some applications where the correlation among the $P$ variables can be safely ignored in the local component  $\boldsymbol{\mathcal{R}}$ (e.g., the traffic data and MODIS data in Section~\ref{sec:experiments}), we directly define $\boldsymbol{K}_{3}^{q}=\frac{1}{\tau^q}\boldsymbol{I}_P$, where $\frac{1}{\tau^q}$ is the variance that can be learned again following Algorithm~\ref{alg_RHyperSample}. Note that one can also add variable-specific variance/precision hyperparameters (e.g., $\tau_p^q$ for $p=1,\ldots,P$) if needed.

\begin{algorithm}[!t]
\caption{Sampling for $\theta_1^{q}$ in $\boldsymbol{\theta}_{\mathcal{R}}^{q}$}
\label{alg_RHyperSample}
\KwIn{$(\theta_1^q)^{(k)}$, $[(\boldsymbol{y}_{\backslash\mathcal{X}}^{q})_{\Omega}]^{(k)}$, $\boldsymbol{r}_{q}^{(k)}$, $(\boldsymbol{L}_{1}^{q})^{(k)}$, $(\boldsymbol{L}_{2}^{q})^{(k)}$, $(\boldsymbol{L}_{3}^{q})^{(k)}$, $\tau^{(k)}$ learned at $k$th MCMC iteration.}
\KwOut{Next $\theta_1^q$, i.e., $(\theta_1^q)^{(k+1)}$.}
Initialize the slice sampling scale $\rho=\log(10)$. \\
Compute the sampling range: $\gamma\sim\text{Uniform}(0,\rho)$, $\theta_{\min}=(\theta_1^q)^{(k)}-\gamma$, $\theta_{\max}=\theta_{\min}+\rho$. \\
Compute the auxiliary variable: $\boldsymbol{G}_{q}=\{(\boldsymbol{L}_{1}^{q})^{(k)}\}^{-1}\boldsymbol{R}_{q}\{((\boldsymbol{L}_{3}^{q})^{(k)})^{\top}\}^{-1}\otimes\{((\boldsymbol{L}_{2}^{q})^{(k)})^{\top}\}^{-1}$, where $\boldsymbol{R}_{q}=\text{reshape}\left(\boldsymbol{r}_{q}^{(k)},[M,TP]\right)$. \\
Draw $\eta\sim\text{Uniform}(0,1)$. \\
\While{True}{
Draw proposal $\theta'\sim\text{Uniform}(\theta_{\min},\theta_{\max})$. \\
Compute $(\boldsymbol{K}_{1}^{q})'$ corresponding to $\theta'$, $(\boldsymbol{L}_{1}^{q})'$ is the Cholesky factor matrix of $(\boldsymbol{K}_{1}^{q})'$. \\
Compute $\boldsymbol{R}_{q}'=(\boldsymbol{L}_{1}^{q})'\boldsymbol{G}_{q}((\boldsymbol{L}_{3}^{q})^{(k)})^{\top}\otimes((\boldsymbol{L}_{2}^{q})^{(k)})^{\top}$, $\boldsymbol{r}_{q}'=\operatorname{vec}(\boldsymbol{R}_{q}')$. \\
\uIf{$\frac{p\left([(\boldsymbol{y}_{\backslash\mathcal{X}}^{q})_{\Omega}]^{(k)}\given\boldsymbol{r}_{q}'\right)p\left(\theta'\right)}{p\left([(\boldsymbol{y}_{\backslash\mathcal{X}}^{q})_{\Omega}]^{(k)}\given\boldsymbol{r}_{q}^{(k)}\right)p\left((\theta_{1}^{q})^{(k)}\right)}>\eta$}
{\Return $(\theta_1^q)^{(k+1)}=\theta'$; \\
\textbf{break};}
\uElseIf{$\theta'<(\theta_{1}^{q})^{(k)}$}
    {$\theta_{\min}=\theta'$;}
\Else{$\theta_{\max}=\theta'$.}
}
\end{algorithm}

\subsubsection{Sampling noise precision $\tau$}

We use a conjugate Gamma prior for $\tau$, and the posterior is still a Gamma distribution $p\left(\tau\given-\right)=\text{Gamma}\left(a^{\ast},b^{\ast}\right)$ with:
\begin{equation}
\begin{aligned}
a^{\ast}&=a_0+\frac{1}{2}|\Omega|, \\
b^{\ast}&=b_0+\frac{1}{2}\left\|\boldsymbol{O}_{1}\operatorname{vec}\left(\boldsymbol{\mathcal{Y}}-\boldsymbol{\mathcal{X}}-\boldsymbol{\mathcal{R}}\right)\right\|_2^2.
\end{aligned}
\end{equation}

\subsection{Model implementation}

For MCMC inference, we perform $K_1$ iterations of the whole sampling process as burn-in and take the following $K_2$ samples for estimation. The predictive distribution over the missing entries $\left\{y_{mtp}^{\ast}:(m,t,p)\in\Omega^c\right\}$ given the observed data points can be approximated by the Monte Carlo estimation:
\begin{equation}
\begin{aligned}
p&\left(y_{mtp}^{\ast}\given\boldsymbol{\mathcal{Y}}_{\Omega},\boldsymbol{\theta}_0\right)\approx \\
&~~~~\frac{1}{K_2}\sum_{k=1}^{K_2}p\left(y_{mtp}^{\ast}\given\boldsymbol{U}^{(k)},\boldsymbol{V}^{(k)},\boldsymbol{W}^{(k)},\boldsymbol{\mathcal{R}}^{(k)},\tau^{(k)}\right),
\end{aligned}
\end{equation}
where $\boldsymbol{\theta}_0=\left\{\mu_{\phi},\tau_{\phi},\mu_{\delta},\tau_{\delta},\mu_{\theta},\tau_{\theta},a_0,b_0\right\}$ represents the set of all parameters for hyper-priors. The variances and credible intervals of the estimations for $y_{mtp}^{\ast}$ can also be obtained using the last $K_2$ samples. We summarize the implementation for BCKL in Algorithm~\ref{alg_BGLK}.

\begin{algorithm}[!t]
\caption{MCMC inference for BCKL}
\label{alg_BGLK}
Initialize $\{\boldsymbol{U},\boldsymbol{V},\boldsymbol{W},\boldsymbol{r}_{q}\}$ as normally distributed random values, $\phi_d=\delta_d=1$, where $d=[1,D]$, $\theta_1^q=\theta_2^q=1$, where $q=[1,Q]$. Set $\mu_{\phi}=\mu_{\delta}=\log{(10)}$, $\mu_{\theta}=\log{(1)}$, $\tau_{\phi}=\tau_{\delta}=\tau_{\theta}=1$, $a_0=b_0=10^{-6}$, $\boldsymbol{\Psi}_0=\boldsymbol{I}_{P}, \nu_0=P$. \\
\For{$k=1:K_{1}+K_{2}$}{
\For{$d=1:D$}{
Draw kernel hyperparameters $\phi_d$; \\
Draw kernel hyperparameters $\delta_d$; \\
Draw $\boldsymbol{u}_{d}\sim\mathcal{N}\left(\boldsymbol{\mu}_{ud}^{\ast},(\boldsymbol{\Lambda}_{ud}^{\ast})^{-1}\right)$; \\
Draw $\boldsymbol{v}_{d}\sim\mathcal{N}\left(\boldsymbol{\mu}_{vd}^{\ast},(\boldsymbol{\Lambda}_{vd}^{\ast})^{-1}\right)$; \\
Draw $\boldsymbol{w}_{d}\sim\mathcal{N}\left(\boldsymbol{\mu}_{wd}^{\ast},(\boldsymbol{\Lambda}_{wd}^{\ast})^{-1}\right)$.
}
Draw hyperparameter $\boldsymbol{\Lambda}_{w}\sim\mathcal{W}\left(\boldsymbol{\Psi}^{\ast},\nu^{\ast}\right)$. \\
\For{$q=1:Q$}{
Draw kernel hyperparameters $\theta_1^q$; \\
Draw kernel hyperparameters $\theta_2^q$; \\
Draw hyperparameter $\boldsymbol{K}_3^q$; \\
Draw sample $\left\{\Tilde{\boldsymbol{r}}_{q},\Tilde{\boldsymbol{y}}_{\backslash\mathcal{X}_1}^{q}\right\}$ from Eq.~\eqref{EqJointsample}; \\
Draw $\boldsymbol{r}_{q}$ from  Eq.~\eqref{Eq_rUpdate}.}
Compute $\boldsymbol{\mathcal{X}}=\sum_{d=1}^{D}\boldsymbol{u}_{d}\circ\boldsymbol{v}_{d}\circ\boldsymbol{w}_{d}$, $\operatorname{vec}\left(\boldsymbol{\mathcal{R}}\right)=\sum_{q=1}^{Q}\boldsymbol{r}_{q}$. \\
Draw model precision $\tau\sim\text{Gamma}(a^{\ast},b^{\ast})$. \\
\uIf{$k>K_1$}{
Compute and collect $\Tilde{\boldsymbol{\mathcal{Y}}}^{(k)}=\boldsymbol{\mathcal{X}}+\boldsymbol{\mathcal{R}}$.}
}
\Return $\left\{\Tilde{\boldsymbol{\mathcal{Y}}}^{(k)}\right\}_{k=K_1+1}^{K_1+K_2}$ to approximate the posterior distributions for the unobserved missing entries.
\end{algorithm}

\subsection{Computational complexity}\label{Sec_Computation}

With Bayesian kernelized tensor factorization, the computational cost for learning the global component and related hyperparameters becomes $\mathcal{O}(M^3+T^3+P^3)$, where $M$, $T$, $P$ are the sizes of the input space. As mentioned, when $M$ or $T$ is large, further computational gains can be achieved with sparse approximation  or using a Gaussian Markov random field (GMRF) with a sparse precision matrix to model spatial/temporal processes. For example, by introducing a sparse approximation with $R$ inducing points, the cost for estimating $\boldsymbol{U}$ is reduced to $\mathcal{O}(R^2M)$. Nevertheless, in a multidimensional setting, even an input space of small size could produce a large dataset. For instance, a 256$\times$256$\times$3 image contains almost 200k pixels. For learning the local component, the computational cost is mainly determined by computing Eq.~\eqref{Eq_rUpdate}, which can be seen as solving a system of  linear equations. Since the coefficient matrix, i.e., $\boldsymbol{K}_{r}^{q}+\boldsymbol{E}'$, is a Kronecker product matrix plus a diagonal matrix, we apply PCG to solve the problem iteratively. The computational cost becomes $\mathcal{O}\left(MTP(M+T+P)\right)$. Particularly, when $\boldsymbol{K}_{3}^{q}$ is assumed to be a diagonal matrix, $\boldsymbol{K}_{r}^{q}$ is highly sparse, and the complexity of calculating Eq.~\eqref{Eq_rUpdate} using PCG further decreases to $\mathcal{O}\left(N_{|\Omega|}|\Omega|\right)$, where $N_{|\Omega|}$ (the average number of the neighbors per data points in the covariance matrix) is close to 1. The overall time cost of BCKL under the general setting can be written as $\mathcal{O}\left(M^2(M+TP)+T^2(T+MP)+P^2(P+MT)\right)$, which is substantially reduced compared to $\mathcal{O}\left(|\Omega|^3\right)$ required by standard GP regression.

\section{Experiments}\label{sec:experiments}

We conduct extensive experiments on both synthetic and real-world datasets, and compare our proposed BCKL framework with several state-of-the-art models. The objective of the synthetic study is to validate the performance of BCKL for modeling nonstationary and nonseparable multidimensional processes. Three types of real-world datasets were used in different tasks/scenarios for assessing the performance of multidimensional data modeling, including imputation for traffic data, completion for satellite land surface temperature data, and color image inpainting as a special tensor completion problem. All codes for reproducing the experiment results are available at \url{https://github.com/mcgill-smart-transport/bckl}.

In terms of evaluation metrics, we consider mean absolute error (MAE) and root mean square error (RMSE) for measuring estimation accuracy:
\begin{equation} \notag
\begin{aligned}
\text{MAE}&=\frac{1}{n}\sum_{i=1}^{n}\left|y_{i}-\hat{y}_{i}\right|, \\
\text{RMSE}&=\sqrt{\frac{1}{n}\sum_{i=1}^{n}\left(y_{i}-\hat{y}_{i}\right)^{2}}, 
\end{aligned}
\end{equation}
where $n$ is the number of estimated data points, $y_i$ and $\hat{y}_{i}$ are the actual value and posterior mean estimation for $i$th test data point, respectively. We also compute continuous rank probability score (CRPS), interval score (INT) \cite{gneiting2007strictly}, and interval coverage (CVG) \cite{heaton2019case} of the 95\% interval estimated on test data to evaluate the performance for uncertainty quantification:
\begin{equation} \notag
\begin{aligned}
\text{CRPS}=&-\frac{1}{n}\sum_{i=1}^{n}\sigma_{i}\Bigg[\frac{1}{\sqrt{\pi}}-2\psi\left(\frac{y_{i}-\hat{y}_{i}}{\sigma_i}\right) \\
&~~~~~~~~~~~~~~~~~~~~-\frac{y_i-\hat{y}_{i}}{\sigma_{i}}\left(2\Phi\left(\frac{y_{i}-\hat{y}_{i}}{\sigma_{i}}\right)-1\right)\Bigg], \\
\text{INT}=&\frac{1}{n}\sum_{i=1}^{n}\left(u_{i}-l_{i}\right)+\frac{2}{\alpha}\left(l_{i}-y_{i}\right)\mathbbm{1}\{y_{i}<l_{i}\} \\
&~~~~~~~~~~~~~~~~~~~~~~~+\frac{2}{\alpha}\left(y_{i}-u_{i}\right)\mathbbm{1}\{y_{i}>u_{i}\}, \\
\text{CVG}=&\frac{1}{n}\sum_{i=1}^{n}\mathbbm{1}\left\{y_{i}\in\left[l_{i},u_{i}\right]\right\},
\end{aligned}
\end{equation}
where $\psi$ and $\Phi$ denote the pdf (probability density function) and cdf (cumulative distribution function) of a standard normal distribution, respectively; $\sigma_{i}$ is the standard deviation (std.) of the estimation values after burn-in for $i$th data, i.e., the std. of $\left\{\Tilde{\boldsymbol{\mathcal{Y}}}^{(k)}\right\}_{k=K_1+1}^{K_1+K_2}$, $\alpha=0.05$, $[l_i,u_i]$ denotes the 95\% central estimation interval for each test data, and $\mathbbm{1}\{\cdot\}$ represents an indicator function that equals 1 if the condition is true and 0 otherwise.

Particularly, for the correlation functions of the local component $\boldsymbol{\mathcal{R}}$, i.e., $k_1^q$ and $k_2^q$, which are assumed to have compact support, we choose the tapering function to be Bohman taper \cite{gneiting2002compactly} in all the following experiments. That is $k_{taper}\left(\Delta;\lambda\right)=\left(1-\frac{\Delta}{\lambda}\right)\cos{\left(\pi\frac{\Delta}{\lambda}\right)}+\frac{1}{\pi}\sin{\left(\pi\frac{\Delta}{\lambda}\right)}$ for $\Delta<\lambda$ and equals to zero for $\Delta\geq\lambda$. $k_1^q$ is then formed by the product of a covariance function $k_0$ and $k_{taper}$: $k_1^q\left(\Delta_1;\theta_1^q\right)=k_0\left(\Delta_1;\theta_1^q\right)k_{taper}\left(\Delta_1;\lambda_1\right)$, and $k_2^q$ is formed similarly.

\subsection{Synthetic study}
\label{Sec_synthetic}

\begin{figure*}[!t]
\centering
\includegraphics[width=0.95\textwidth]{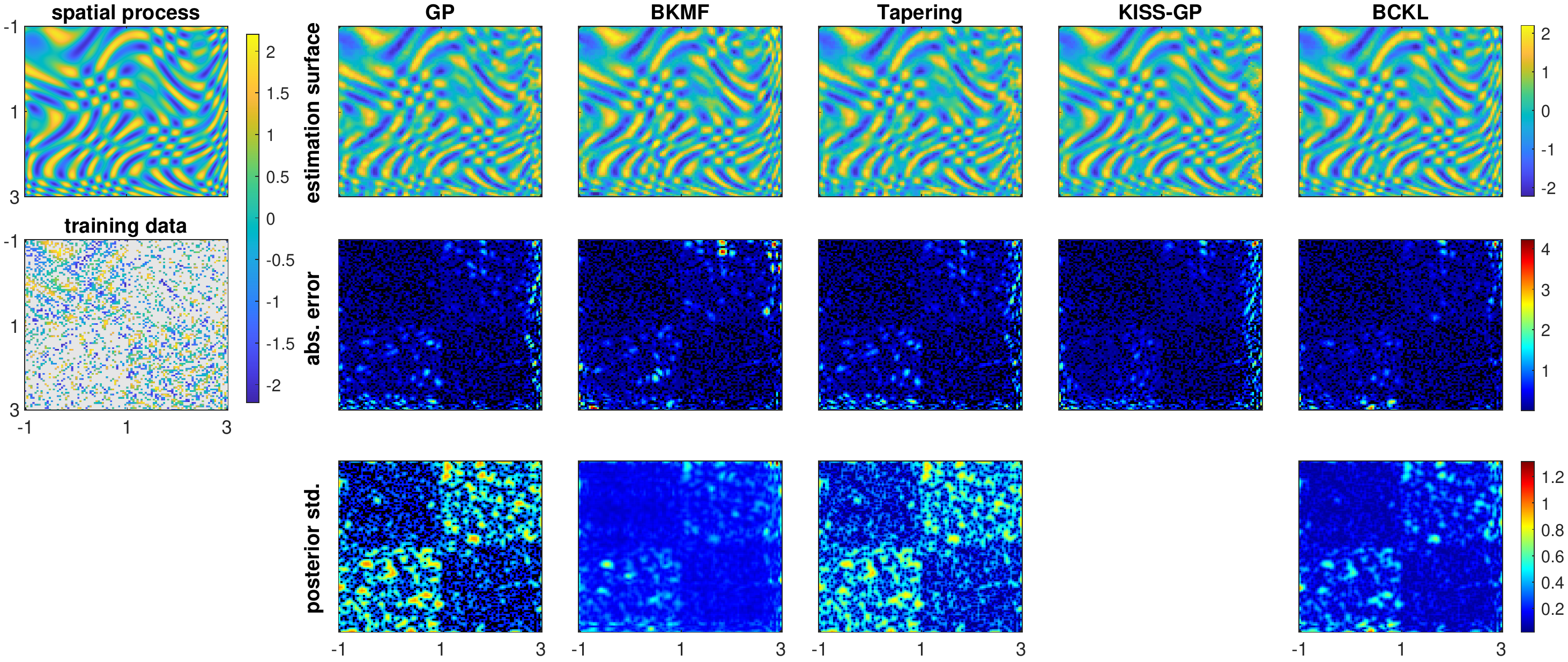}
\caption{Illustration for model performance on a nonstationary and nonseparable spatial process, where the gray parts in training data denote missing points. The figure plots the estimation surface, the absolute estimation error on the test data, and the estimation std., obtained by different models. Note that KISS-GP does not provide uncertainty estimations.}
\label{fig_Intro}
\end{figure*}

We first generate a nonstationary and nonseparable spatial random field in $\mathbb{R}^2$ to test the effectiveness of the proposed framework. A $100\times 100$ 2D spatial process $\boldsymbol{Y}$ is created in a $[-1.0,3.0]\times[-1.0,3.0]$ square, following:
\begin{equation}\notag
\begin{aligned}
\boldsymbol{Y}(s_1,s_2)=&\cos{\left\{4\left[f_1(s_1)+f_2(s_2)\right]\right\}}\\
&+\sin{\left\{4\left[f_1(s_2)-f_2(s_1)\right]\right\}},
\end{aligned}
\end{equation}
where $s_1,s_2\in[-1.0,3.0]$ are the coordinates of the first and second dimensions, respectively; $f_1(s)=s\left(\sin{(2s)}+2\right)$, and $f_2(s)=0.2s\sqrt{99(s+1)+4}$. An i.i.d. noise with a variance of 0.01 is further added for the experiment. To generate training and test datasets, we divide the space into four parts, then set 60\% random missing for the two parts on the diagonal, and 80\% random missing for the other two. The spatial processes are shown in Fig.~\ref{fig_Intro}.

In this 2D case, the global tensor factorization and local 3D component in the proposed BCKL framework become a kernelized MF and a 2D local process, respectively. Specifically, we use a squared exponential (SE) kernel to build the covariance matrices for the global latent factors of both input dimensions. For the local component, the kernel functions of $k_0$ in $k_1^q$ and $k_2^q$ are still SE kernels, and we apply Bohman taper with range $\{\lambda_1,\lambda_2\}$ as $\{10,10\}$. We set rank $D=10$, and the number of local components $Q=2$. For model inference, we run in a total of 1500 MCMC iterations, where the first 1000 runs were burn-in. Several models are compared in this synthetic spatial field. The baseline methods include a stationary GP with SE kernels fitting on the vectorized training data, a low-rank model BKMF (Bayesian kernelized MF) with $D=20$ \cite{lei2022bayesian}, a local GP approach derived from covariance tapering \cite{furrer2006covariance}, and a kernel interpolation model KISS-GP \cite{wilson2015kernel}. We form the tapering kernel as the sum of two local product kernels, with the same setting as the local component in BCKL. For KISS-GP, we apply a mixture spectral kernel for each dimension with 5 frequency bases and tune the model using the GPML toolbox\footnote{\url{http://gaussianprocess.org/gpml/code/matlab/doc/}}.

Fig.~\ref{fig_Intro} also compares the imputation surfaces of different models, illustrating the shortcomings of both low-rank and GP models and moreover how the combination of both can be leveraged for improved estimation. One can observe that BKMF is able to approximate the underlying trend of the data, but the residual is still locally correlated. The three GP-based models, i.e., stationary GP, tapering, and KISS-GP, can fit local variations but lose the nonstationary global structure, which particularly leads to high estimation errors when the data length scale rapidly changes from long to short (e.g., the bottom and right edges of the spatial field). The proposed BCKL model, on the other hand, can learn both constructional dependencies and local changes in the data, which solves the issues of BKMF and GP models. The quantified imputation performance of different models is compared in Table~\ref{Synthetic-comTable}. The proposed BCKL clearly achieves the best estimation results with the lowest posterior mean estimation errors and uncertainty scores (CRPS and INT).

\begin{table}[!t]
\footnotesize
  \centering
  \caption{Synthetic imputation performance.}
    \begin{tabular}{l|rrrrr}
    \toprule
    Metrics & GP & BKMF & Tapering & KISS-GP & BCKL  \\
    \midrule
    MAE & 0.28 & 0.28 & 0.29 & 0.26 & \textbf{0.21} \\
    RMSE & 0.44 & 0.47 & 0.44 & 0.43 & \textbf{0.34} \\
    CRPS & 0.21 & 0.21 & 0.21 & - & \textbf{0.15}  \\
    INT & 2.49 & 2.81 & 2.45 & - & \textbf{1.58} \\
    CVG & \textbf{0.94} & 0.91 & 0.93 & - & 0.93 \\
    \bottomrule
    \multicolumn{6}{l}{{Best results are highlighted in bold fonts.}}
    \end{tabular}
  \label{Synthetic-comTable}
\end{table}

\subsection{Traffic data imputation}

\subsubsection{Datasets}

In this section, we perform spatiotemporal modeling on two traffic speed datasets:

\begin{itemize}
\item[(S):] Seattle traffic speed dataset\footnote{\url{https://github.com/zhiyongc/Seattle-Loop-Data}}. This dataset contains traffic speed collected from 323 loop detectors on the Seattle freeway in 2015 with a 5-minutes interval (288 time points per day). We use the data of 30 days (from Jan 1st to Jan 30th) in the experiments and organize it into a $\textit{location}\times\textit{time points per day}\times\textit{day}$ tensor with the size of $323\times 288\times 30$ ($M\times T\times P$).

\item[(P):] PeMS-Bay traffic speed dataset\footnote{\url{https://github.com/liyaguang/DCRNN}}. This dataset consists of traffic speed observations collected from 325 loop detectors in the Bay Area, CA. We select a 2-month subset (from Feb 1st to March 31th, 2017) of 319 sensors for the experiments, and aggregate the records by 10-minutes windows (144 time points per day). The applied dataset is also represented as a $\textit{location}\times\textit{time points per day}\times\textit{day}$ tensor of size $319\times 144\times 59$. 
\end{itemize}

The global factors $\boldsymbol{u}_{d}$ here correspond to the spatial sensor dimension and the factors $\boldsymbol{v}_{d}$ represent the temporal evolution of the data in one day. We set the kernel priors for the global latent factors $\{\boldsymbol{u}_{d},\boldsymbol{v}_{d}\}$ following \cite{lei2022bayesian}. For the covariance function of $\boldsymbol{u}_{d}$, i.e., $k_u\left(\Delta_u;\phi_d\right)$, we use graph diffusion kernel and regularized Laplacian kernel \cite{smola2003kernels} for the (S) and (P) dataset, respectively. When constructing the kernel matrices, a sensor network adjacency matrix $\boldsymbol{A}\in\mathbb{R}^{M\times M}$ is firstly built following $\exp\left\{-{\Delta_u^2}/{\phi_d^2}\right\}$ to capture the edge weight between each pair of the input sensors \cite{li2018dcrnn_traffic}, where $\Delta_u$ denotes the pairwise shortest road network distance between sensors, $\phi_d$ is the length-scale hyperparameter. A normalized Laplacian matrix $\Tilde{\boldsymbol{L}}$ \cite{smola2003kernels} is then computed based on $\boldsymbol{A}-\boldsymbol{I}_{M}$, and the covariance matrix $\boldsymbol{K}_{u}^{d}$ is  constructed by $\boldsymbol{K}_{u}^{d}=\exp{(-\Tilde{\boldsymbol{L}})}$ and $\boldsymbol{K}_{u}^{d}=(\boldsymbol{I}_{M}+\Tilde{\boldsymbol{L}})^{-1}$ for data (S) and (P), respectively. For $k_v\left(\Delta_v;\delta_d\right)$, we apply a Matern 3/2 kernel \cite{williams2006gaussian} for both datasets, with $\Delta_v$ denoting the distance between time points and $\delta_d$ being the length-scale hyperparameter.

For the local component $\boldsymbol{\mathcal{R}}$, as mentioned, the correlation functions for the first two dimensions are constructed through $k_1^q\left(\Delta_1;\theta_1^q\right)=k_0\left(\Delta_1;\theta_1^q\right)k_{taper}\left(\Delta_1;\lambda_1\right)$ and $k_2^q\left(\Delta_2;\theta_2^q\right)=k_0\left(\Delta_2;\theta_2^q\right)k_{taper}\left(\Delta_2;\lambda_2\right)$, respectively, where $k_{taper}$ is Bohman function. We select the kernel functions of $k_0$ in $k_1^q$ and $k_2^q$ to be the same as $k_u$ and $k_v$, respectively, with $\theta_1^q$ and $\theta_2^q$ denoting the length-scale hyperparameters. The range parameters in $k_{taper}$, i.e., $\{\lambda_1,\lambda_2\}$, are set as $\{200,20\}$ and $\{4,10\}$ for (S) and (P), respectively. As we do not expect the residual process to be correlated among different days, we let $\boldsymbol{K}_3^q=\frac{1}{\tau^q}\boldsymbol{I}_P$.

\subsubsection{Experimental settings}

\textbf{Missing scenarios.}
To evaluate the performance of our model, we test the imputation task and consider three types of missing scenarios: random missing (RM), nonrandom missing (NM), i.e., random whole-day tube missing by masking a tube $\boldsymbol{\mathcal{X}}(m,:,p)$ (see \cite{chen2021bayesian}), and single time point blackout missing (SBM) by randomly masking a tube $\boldsymbol{\mathcal{X}}(:,t,p)$. In practice, NM refers to the scenario where a certain amount of sensors are not working (e.g., sensor failure) each day, and SBM refers to the scenario where all sensors are not working over certain time points (e.g., communication and power failures). We set the percentage of missing values as 30\% and 70\% in both RM and NM scenarios, and as 50\% in SBM scenario.

\textbf{Baselines.}
We compare the proposed BCKL framework with the following baseline models:
\begin{itemize}
    \item Bayesian probabilistic tensor factorization (BPTF): a pure low-rank Bayesian CP factorization model without GP priors on the latent factors, which is similar to Bayesian Gaussian CP decomposition (BGCP) \cite{chen2019bayesian}. 
    \item Bayesian kernelized tensor factorization (BKTF): a third-order tensor extension of BKMF \cite{lei2022bayesian} using CP factorization, which is a special case of BCKL without the local component $\boldsymbol{\mathcal{R}}$. 
    \item Tapering method: a GP regression utilizing the sum of a product of compactly supported covariance functions constructed by covariance tapering \cite{luttinen2012efficient}. This is a special case of BCKL without the global component $\boldsymbol{\mathcal{X}}$. 
    \item Kernel Interpolation for Scalable Structured GP (KISS-GP) \cite{wilson2015kernel}: a state-of-the-art GP model for large-scale multidimensional datasets that contain missing values.
\end{itemize}
For BCKL, the CP rank $D$ for approximating $\boldsymbol{\mathcal{X}}$ is set as 20 for 30\% RM and 30\% NM, 10 for 70\% RM and 70\% NM, and 15 for 50\% SBM. In all the scenarios, we test using only one and the sum of two local variables to estimate $\boldsymbol{\mathcal{R}}$, i.e., $Q=\{1,2\}$, denoting by BCKL-I and BCKL-II, respectively. 
For BPTF, we use the same rank assumptions as BCKL for each scenario. All the settings of BKTF, including rank $D$, kernel functions for latent factors $k_u$ and $k_v$, and other model parameters, are the same as the global component $\boldsymbol{\mathcal{X}}$ in BCKL. For the tapering method, we use the sum of two local product kernel functions ($Q=2$). For KISS-GP, similar to the synthetic study, each input dimension is modeled through a spectral mixture kernel and learned by the GPML toolbox. Specifically, we trained the kernel with $\{5,10,15\}$ spectral components for each scenario, and report the best results. For MCMC, we run 600 iterations as burn-in and take the next 400  for estimation for all Bayesian models.

\subsubsection{Results}

\begin{table*}[!t]
\footnotesize
\setlength{\tabcolsep}{1.5mm}
  \centering
  \caption{Imputation performance on traffic datasets.}
    \begin{tabular}{llc|cccccc}
    \toprule
    Data & Scenarios & Metrics & BPTF & BKTF & Tapering & KISS-GP & BCKL-I & BCKL-II \\
    \midrule
    \multirow{10}{*}{(S)} & \multirow{2}{*}{30\% RM} & MAE/RMSE & 3.14/5.10 & 3.15/5.11 & 2.14/3.21 & 2.96/4.55 & 2.19/3.26 & \textbf{2.11}/\textbf{3.16}  \\
    & & CRPS/INT/CVG & 2.51/33.34/\textbf{0.95} & 2.52/33.28/\textbf{0.95} & 1.65/19.71/0.94 & - & 1.68/20.06/0.94 & \textbf{1.62}/\textbf{19.45}/\textbf{0.95} \\
    \cmidrule(r){2-9}
    & \multirow{2}{*}{70\% RM} & MAE/RMSE & 3.42/5.53 & 3.42/5.53 & 2.66/4.06 & 3.17/4.94 & 2.48/3.77 & \textbf{2.43}/\textbf{3.70}  \\
    & & CRPS/INT/CVG & 2.73/36.23/0.94 & 2.74/36.20/\textbf{0.95} & 2.06/25.21/0.94 & - & 1.92/23.47/0.94 & \textbf{1.88}/\textbf{22.98}/0.94  \\
    \cmidrule(r){2-9}
    & \multirow{2}{*}{30\% NM} & MAE/RMSE & 3.35/5.49 & 3.32/5.52 & 5.09/7.86 & 3.44/5.38 & 2.88/4.55 & \textbf{2.85}/\textbf{4.51}  \\
    & & CRPS/INT/CVG & 2.65/36.12/\textbf{0.94} & 2.66/36.14/\textbf{0.94} & 3.86/52.98/0.92 & - & 2.25/28.75/\textbf{0.94} & \textbf{2.23}/\textbf{28.61}/\textbf{0.94} \\
    \cmidrule(r){2-9}
    & \multirow{2}{*}{70\% NM} & MAE/RMSE & 3.71/6.24 & 3.67/6.04 & 6.29/9.92 & 4.07/6.54 & \textbf{3.40}/\textbf{5.56} & 3.41/5.61  \\
    & & CRPS/INT/CVG & 3.08/46.13/0.93 & 2.94/40.45/0.93 & 4.80/71.06/0.93 & - & \textbf{2.67}/\textbf{36.24}/\textbf{0.94} & 2.74/37.05/\textbf{0.94}  \\
    \cmidrule(r){2-9}
    & \multirow{2}{*}{50\% SBM} & MAE/RMSE & 3.35/5.41 & 3.29/5.37 & 2.46/3.70 & 3.13/4.89 & 2.45/3.67 & \textbf{2.41}/\textbf{3.63} \\
    & & CRPS/INT/CVG & 2.66/35.35/0.94 & 2.63/34.84/\textbf{0.95} & 1.89/22.89/0.94 & - & 1.88/22.75/0.93 & \textbf{1.85}/\textbf{22.51}/0.94 \\
    \midrule
    \multirow{10}{*}{(P)} & \multirow{2}{*}{30\% RM} & MAE/RMSE & 2.22/4.10 & 2.19/4.08 & 0.94/1.74 & 1.74/3.24 & 0.90/1.66 & \textbf{0.89}/\textbf{1.63}  \\
    & & CRPS/INT/CVG & 1.92/27.90/\textbf{0.95} & 1.91/27.77/\textbf{0.95} & 0.79/11.45/\textbf{0.95} & - & 0.76/10.99/\textbf{0.95} & \textbf{0.75}/\textbf{10.80}/\textbf{0.95} \\
    \cmidrule(r){2-9}
    & \multirow{2}{*}{70\% RM} & MAE/RMSE & 2.42/4.49 & 2.43/4.51 & 1.72/3.45 & 2.72/5.08 & 1.34/2.64 & \textbf{1.33}/\textbf{2.62}  \\
    & & CRPS/INT/CVG & 2.10/30.97/\textbf{0.95} & 2.12/30.83/\textbf{0.95} & 1.47/21.80/\textbf{0.95} & - & 1.17/17.42/\textbf{0.95} & \textbf{1.16}/\textbf{17.24}/\textbf{0.95} \\
    \cmidrule(r){2-9}
    & \multirow{2}{*}{30\% NM} & MAE/RMSE & 2.48/4.80 & 2.46/4.81 & 5.27/8.77 & 3.47/6.13 & \textbf{2.25}/\textbf{3.98} & 2.27/4.01  \\
    & & CRPS/INT/CVG & 2.12/33.09/0.93 & 2.11/32.77/\textbf{0.94} & 4.11/65.00/0.93 & - & \textbf{1.86}/\textbf{26.63}/\textbf{0.94} & 1.88/26.82/\textbf{0.94} \\
    \cmidrule(r){2-9}
    & \multirow{2}{*}{70\% NM} & MAE/RMSE & 2.81/5.42 & 2.77/5.35 & 5.56/9.20 & 4.44/7.65 & \textbf{2.73}/\textbf{4.76} & 2.74/4.79  \\
    & & CRPS/INT/CVG & 2.32/35.93/0.93 & 2.32/35.76/0.93 & 4.32/68.43/0.93 & - & \textbf{2.24}/\textbf{32.19}/0.93 & 2.25/32.39/\textbf{0.94} \\
    \cmidrule(r){2-9}
    & \multirow{2}{*}{50\% SBM} & MAE/RMSE & 2.48/4.91 & 2.32/4.31 & 1.19/2.35 & 2.73/5.03 & 1.08/2.06 & \textbf{1.05}/\textbf{2.03}   \\
    & & CRPS/INT/CVG & 2.07/30.57/0.94 & 2.01/29.34/\textbf{0.95} & 1.01/15.04/\textbf{0.95} & - & 0.92/13.61/\textbf{0.95} & \textbf{0.91}/\textbf{13.52}/\textbf{0.95} \\
    \bottomrule
    \multicolumn{4}{l}{{Best results are highlighted in bold fonts.}}
    \end{tabular}
  \label{TrafficData-comTable}
\end{table*}

The imputation performance of BCKL and baseline methods for the two traffic speed datasets are summarized in Table~\ref{TrafficData-comTable}. Evidently, the proposed BCKL model consistently achieves the best performance in all cases. In most cases, BCKL-II ($Q=2$) outperforms BCKL-I ($Q=1$), showing the importance of additive kernels in capturing nonseparable correlations. We see that BPTF and BKTF have similar estimation errors when the data is missing randomly, but BKTF outperforms BPTF in NM and SBM scenarios due to the incorporation of GP priors over space and time. Benefiting from the global patterns learned by the low-rank factorization, BKTF and BPTF offer competitive mean estimation accuracy, particularly in NM scenarios, but uncertainty quantification is not as desired. Even though the CRPS/INT/CVG for 95\% intervals of BKTF are marginally better than BPTF, the results are still less than satisfactory. By contrast, the GP covariance tapering method can obtain superior uncertainty results and comparable MAE/RMSE, especially in RM and SBM scenarios, where short-scale variations can be learned from the observations and used to impute or interpolate for the missing values. However, the tapering model fails in NM scenarios, since the local dynamics cannot be effectively leveraged for the whole-day missing cases. Clearly, the BCKL framework possesses the advantages of both the low-rank kernelized factorization and the local GP processes: it is able to learn both the global long-range dependencies (which is important for NM) and the local short-scale spatiotemporal correlations (which is important for RM/SBM) of the data and provide high-quality estimation intervals.

\begin{figure*}[!t]
\centering
\includegraphics[width=1\textwidth]{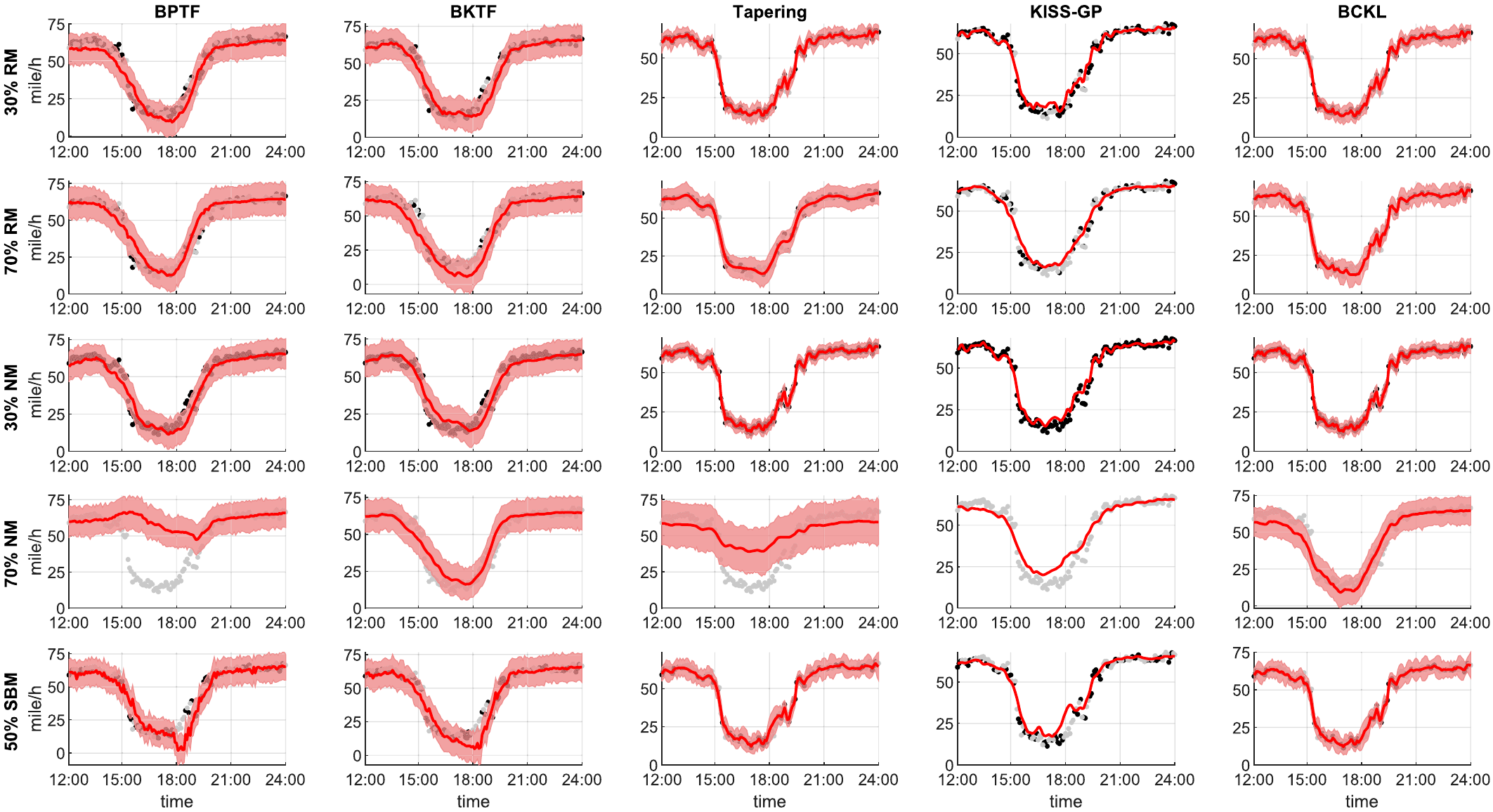}
\caption{Comparison of imputation results on data (S) at sensor \#200 and 15th day. Black and gray dots correspond to the observed and missing data, respectively. Red curves show the posterior mean estimated from different methods, and the shaded areas are 95\% credible intervals. The results of BCKL are from BCKL-I/II with better RMSE and CRPS. Note that KISS-GP does not calculate uncertainty.}
\label{fig_SeDataImpuResults}
\end{figure*}

\begin{figure}[!t]
\centering
\includegraphics[width=0.5\textwidth]{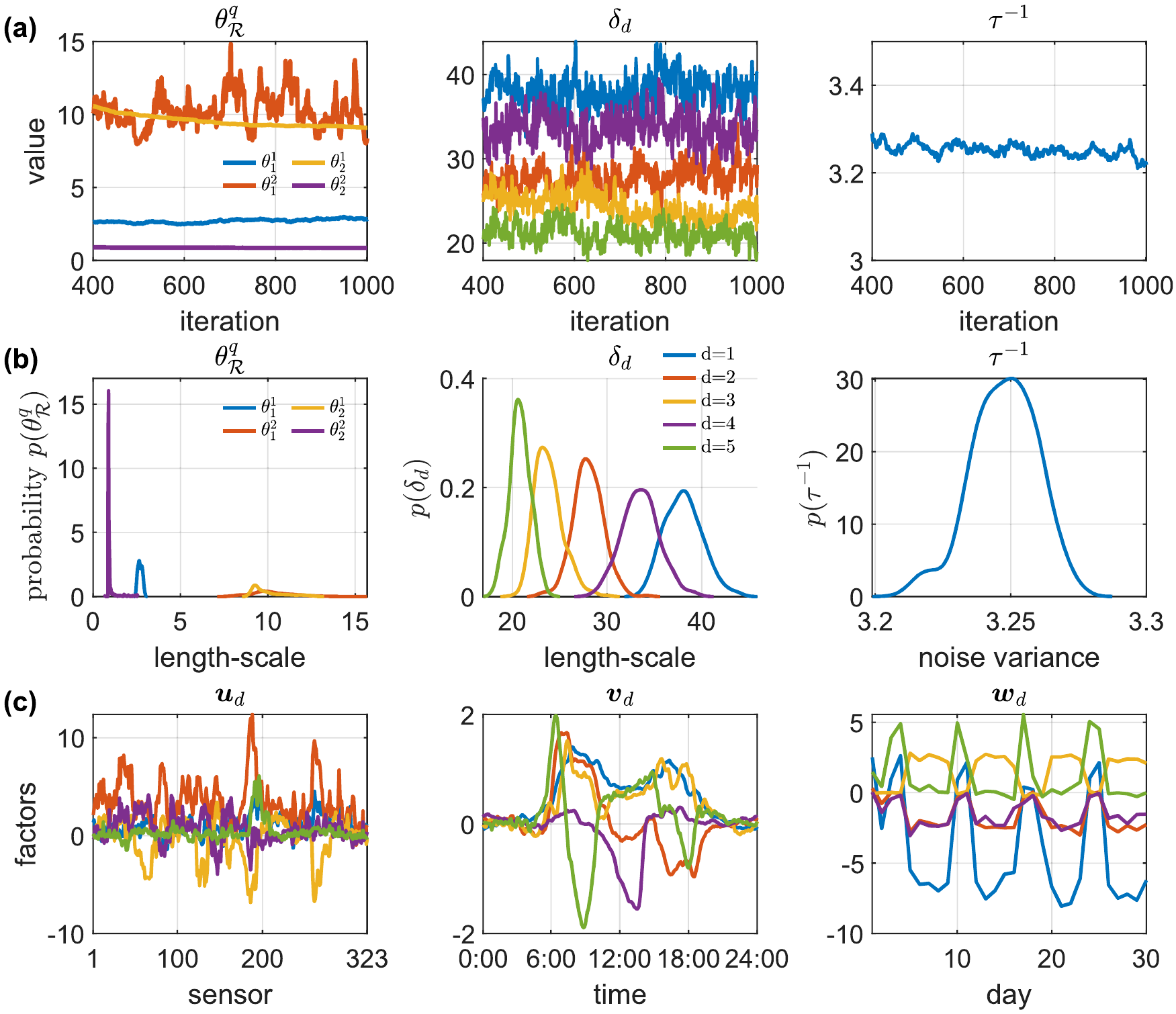}
\caption{Examples of the kernel hyperparameters and model parameters learned by BCKL for (S) data 30\% RM. (a) trace plots of the last 600 MCMC samples; (b) posterior probability density functions; (c) latent factors. The legends for the plots of $\delta_d$ and $\boldsymbol{u}_{d},\boldsymbol{v}_{d},\boldsymbol{w}_{d}$ are the same.}
\label{fig_SeDataResultsOther}
\end{figure}

\begin{figure}[!t]
\centering
\centering
\includegraphics[width=0.49\textwidth]{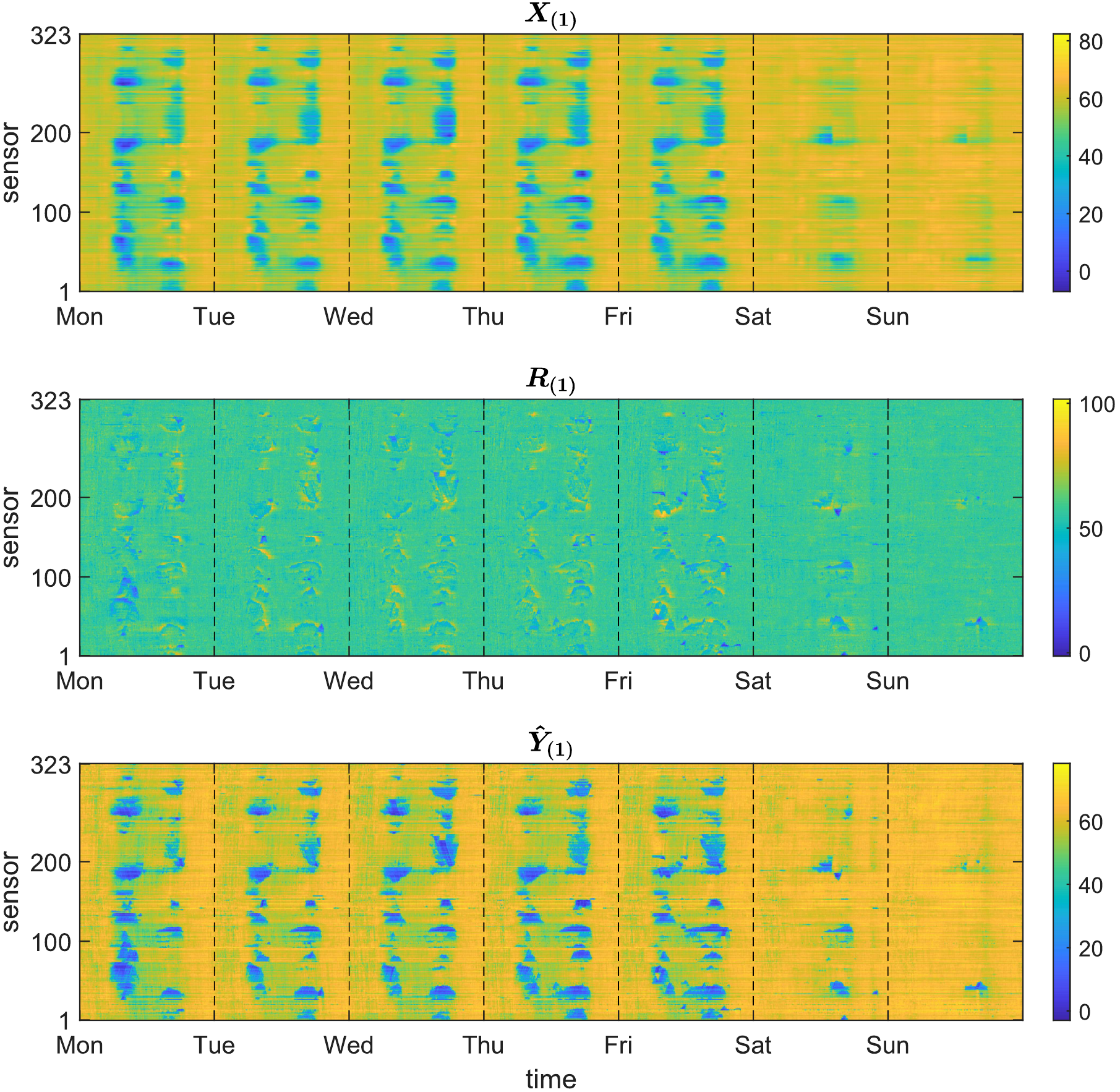}
\caption{Illustration of the global component $\boldsymbol{\mathcal{X}}$, local component $\boldsymbol{\mathcal{R}}$, and output data $\boldsymbol{\mathcal{Y}}$ estimated by BCKL for (S) data 30\% RM. The figure shows mode-1 unfolding matrices of the three estimated tensors for one week.}
\label{fig_SeResult}
\end{figure}

Fig.~\ref{fig_SeDataImpuResults} shows an example of the estimation results on dataset (S). It is clear that for the baseline models, low-rank BKTF performs better in NM scenarios compared with others, while the local GP tapering approach is better at handling RM and SBM scenarios. On the other hand, we see that the proposed BCKL model can take advantage of both the global and the local consistencies in all situations. Fig.~\ref{fig_SeDataResultsOther} illustrates the trace plots, posterior distributions of the kernel hyperparameters $\{\delta_d,\theta_1^q,\theta_2^q\}$ and model noise variance $\tau^{-1}$, along with the global latent factors $\{\boldsymbol{u}_{d},\boldsymbol{v}_{d},\boldsymbol{w}_{d}:d=[1:5]\}$ learned by BCKL for data (S) 30\% RM. The trace plots display converged MCMC chains with good mixing, implying that BCKL learned a stable posterior joint distribution with fast convergence rates. Comparing the first two panels in (b) which show $\delta_d$ and $\theta_2^q$ (i.e., length-scales of the global factors and the local components for the \textit{time of day} dimension), respectively, we can see that the values of $\delta_d$ are much larger than $\theta_2^q$. This suggests that the global and local variables can separately capture the long-range flat variations and the short-scale rapid variations of the data, which is consistent with the overall model assumptions. From the latent factors that correspond to the \textit{time of day} and \textit{day} dimension, i.e., $\boldsymbol{v}_{d}$ and $\boldsymbol{w}_{d}$, one can clearly observe the daily morning/evening peaks and weekly weekday/weekend patterns, respectively. Fig.~\ref{fig_SeResult} gives examples of BCKL estimated global and local components for data (S) imputation, from which we can intuitively see the long-range periodic patterns and short-term local correlated structures of $\boldsymbol{\mathcal{X}}$ and $\boldsymbol{\mathcal{R}}$, respectively.

\subsection{MODIS satellite temperature  completion/kriging}

\subsubsection{Datasets}

Through satellite imaging, massive high-resolution spatiotemporal datasets can be collected. A common issue of satellite image data is the corruption/absense of large regions obstructed by clouds. Here we analyze daily latticed land surface temperature (LST) data in 2020 collected from the Terra platform onboard the MODIS (Moderate Resolution Imaging Spectroradiometer) satellite\footnote{\url{https://modis.gsfc.nasa.gov/data/}}. This type of data has been widely used in the literature as a kriging task to benchmark scalable GP models \cite{heaton2019case}. For our experiment, the spatial resolution is 0.05 degrees latitude/longitude, and we select a $100\times 200$ grids space with the latitude and longitude ranging from 35 to 39.95 and from -114.95 to -105, respectively. We conduct kriging experiments on three subsets with different size w.r.t. number of days: Aug for 1mon, \{Jul, Aug\} for 2mon, and \{Jul, Aug, Sep\} for 3mon, which are represented as a $\textit{latitude}\times \textit{longitude}\times \textit{day}$ ($M\times N\times P$) tensors of size $100\times 200\times 31$, $100\times 200\times 62$, $100\times 200\times 92$, respectively. The amounts of missing values are 6.34\%, 10.99\%, and 12.17\% for 1mon, 2mon, and 3mon, respectively. The data unit is transformed to kelvin during the analysis. It should be noted that we no longer have a temporal dimension in this experiment. Instead, we separate longitude and latitude as two different spatial dimensions, and thus we still have a third-order tensor structure.

We use Matern 3/2 kernels to construct the covariance functions for both $\boldsymbol{u}_{d}$ and $\boldsymbol{v}_{d}$. The local correlation functions $\{k_1^q,k_2^q\}$ are built in a similar way as the traffic datasets, where all $k_0$ are Matern 3/2 kernel and the range parameters in $k_{taper}$ are set as $\{30,30\}$ for all the three datasets.

\subsubsection{Experimental settings}

\textbf{Missing scenarios.}
To create realistic missing patterns, we utilize the missing patterns of MODIS LST data in Jul 2021 from the same spatial region to generate missing scenarios for the three applied datasets. In detail, for data on each day (each $100\times 200$ slice matrix of the data tensor), we mask those indices that are missing on the same day of the month in Jul 2021. Then, those masked but observed indices are used as test data for evaluation. The overall sizes of test data in 1mon, 2mon, and 3mon are 14.80\%, 14.33\%, and 14.93\%, respectively. 

\noindent\textbf{Baselines.} We still compare BCKL with BPTF, BKTF, Tapering, and KISS-GP for this experiment; and still run 600 MCMC iterations for burn-in and take the following 400 samples for estimation. In BCKL, the rank $D$ is set to 70, and $Q$ is set to 2 for all the 1-3mon datasets. Again, as we do not expect the local variation to be correlated across different days, we set $\boldsymbol{K}_3^q=\frac{1}{\tau^q}\boldsymbol{I}_P$, instead of generating a full $P\times P$ covariance from a Wishart distribution. The rank settings for BPTF and BKTF, the kernel assumptions for BKTF, and the construction of covariance functions for the tapering method is the same as the corresponding settings in BCKL. For the tapering approach, in this case, we introduce a second-order polynomial trend surface as the mean function, i.e., $\boldsymbol{\mu}(LAT, LONG, DAY;\boldsymbol{\beta})=\beta_0^{DAY}+\beta_1LONG+\beta_2LAT+\beta_3LONG^2+\beta_4LAT^2+\beta_5LONG\times LAT$, where we consider longitude (LONG) and latitude (LAT) as continuous covariates and (DAY) as a categorical covariate. The tuning processes and settings for KISS-GP are the same as the traffic datasets.

\subsubsection{Results}

\begin{table*}[!t]
\footnotesize
  \centering
  \caption{Completion performance on MODIS LST datasets.}
    \begin{tabular}{lc|ccccc|c}
    \toprule
    Data & Metrics & BPTF & BKTF & Tapering & KISS-GP & BCKL & BCKL-Aug \\
    \midrule
    \multirow{2}{*}{1mon} & MAE/RMSE & 2.13/2.90 & 2.17/2.94 & 3.59/4.98 & 3.68/5.15 & \textbf{1.90}/\textbf{2.67} & 1.90/2.67  \\
    & CRPS/INT/CVG & 1.57/17.18/0.90 & 1.59/15.71/\textbf{0.92} & 2.58/24.64/0.90 & - & \textbf{1.40}/\textbf{15.09}/\textbf{0.92} & 1.40/15.09/0.92 \\
    \midrule
    \multirow{2}{*}{2mon} & MAE/RMSE & 2.32/3.15 & 2.37/3.20 & 3.75/5.20 & 3.27/4.30 & \textbf{1.97}/\textbf{2.76} & 1.85/2.60  \\
    & CRPS/INT/CVG & 1.71/18.78/0.89 & 1.73/16.95/\textbf{0.94} & 2.69/25.48/0.90 & - & \textbf{1.45}/\textbf{15.34}/0.93 &  1.37/14.28/0.94 \\
    \midrule
    \multirow{2}{*}{3mon} & MAE/RMSE & 2.01/2.77 & 2.13/2.90 & 3.49/4.86 & 2.95/3.82 & \textbf{1.76}/\textbf{2.51} & 1.62/2.30  \\
    & CRPS/INT/CVG & 1.49/17.25/0.88 & 1.57/15.59/\textbf{0.94} & 2.51/24.24/0.90 & - & \textbf{1.31}/\textbf{14.10}/0.93 & 1.20/12.62/0.95 \\
    \bottomrule
    \multicolumn{4}{l}{Best results are highlighted in bold fonts.}
    \end{tabular}
  \label{MODIS-comTable}
\end{table*}

The interpolation results of different approaches on the three MODIS LST datasets are listed in Table~\ref{MODIS-comTable}. BCKL evidently outperforms other baseline methods for all three subsets with the highest estimation accuracy and the best uncertainty quality. The last column of Table~\ref{MODIS-comTable} gives the estimation performance of BCKL on the data of August when applying different datasets. This result shows that the completion of August can benefit from accessing the corrupted data in July and September, suggesting that having more days of data could enhance the estimation of the local component by leveraging the correlations among different days. For the tapering method, even with a second-order polynomial mean surface, it is still difficult to perform such completion tasks when large chunks of data are missing due to the presence of clouds. This is mainly because that $\boldsymbol{K}_3^q=\frac{1}{\tau^q}\boldsymbol{I}_P$ implies that images from different days follow independent spatial processes. It is clearly challenging to fill a large missing block relying only on local dependencies in GP, since long-range and cross-day dependencies encoded in the global component play a key role in reconstructing the data.

\begin{figure}[!t]
\centering
\includegraphics[width=0.49\textwidth]{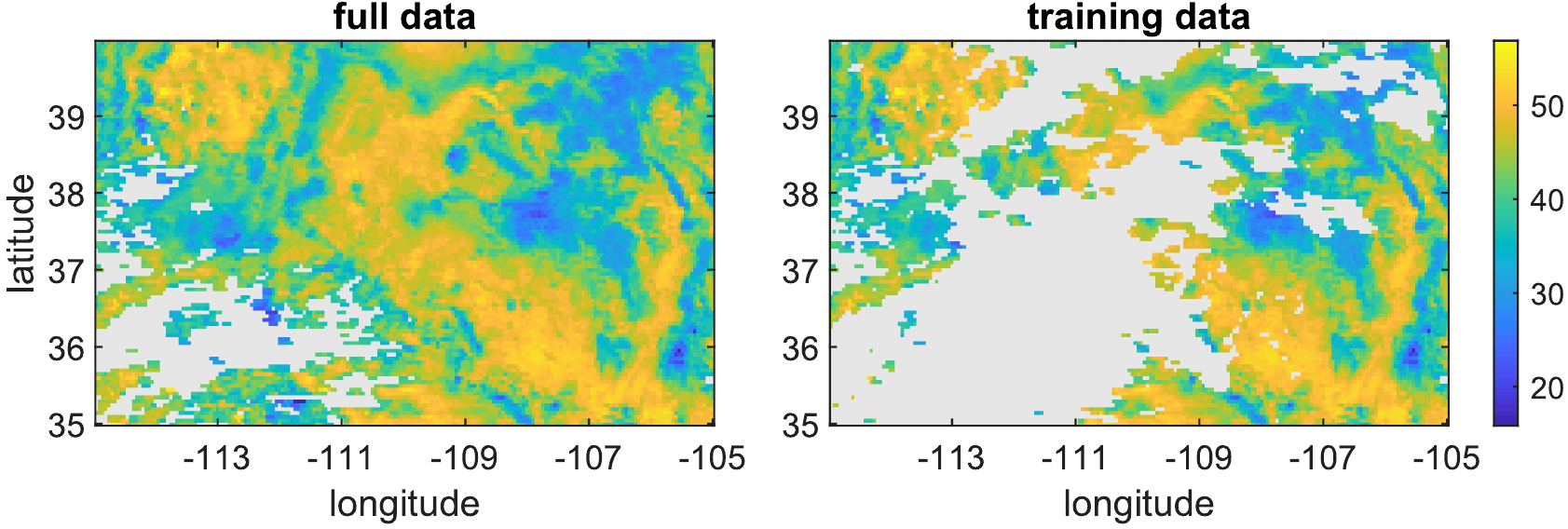}
\caption{Full and training data of MODIS LST on Aug 13th. The gray area means data is missing. The difference between these is used as test data for the kriging task.}
\label{fig_MODIStrain}
\end{figure}

\begin{figure*}[!t]
\centering
\includegraphics[width=1\textwidth]{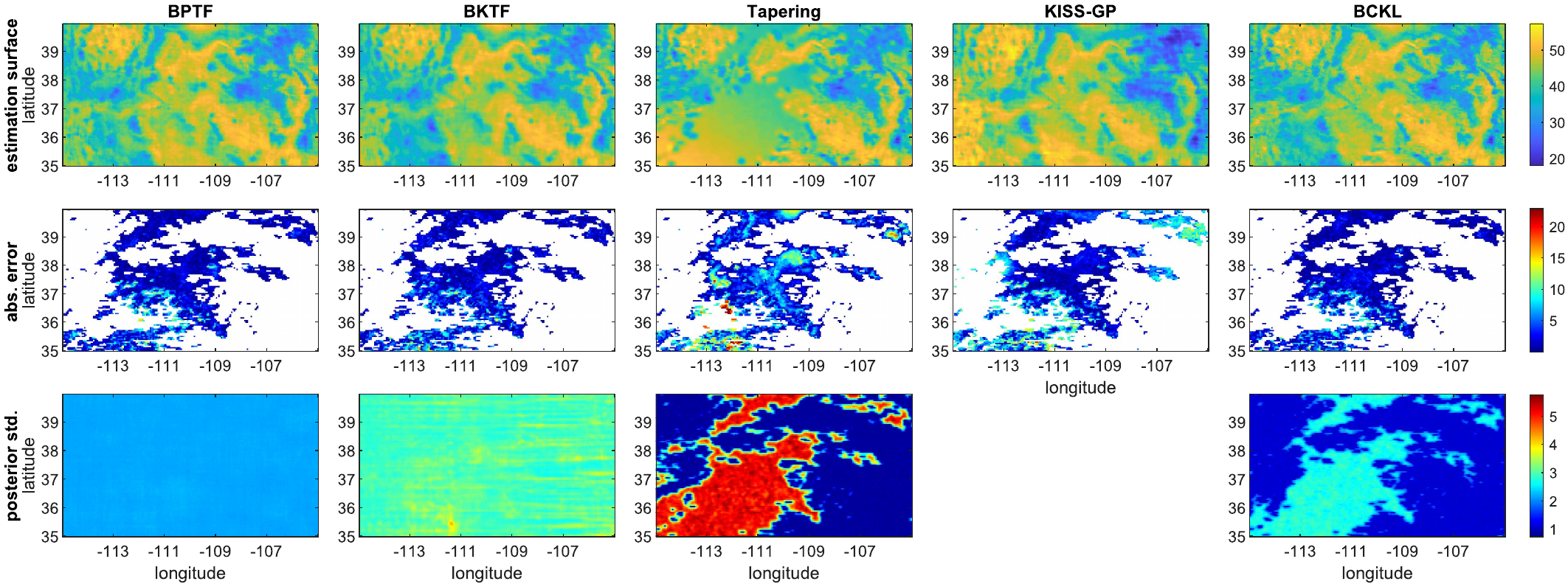}
\caption{Completion results of different methods for MODIS LST on Aug 13th, including the estimation surfaces, the absolute estimation errors for the test data, and the posterior std. of the estimations computed from the last $K_2$ samples.}
\label{fig_MODISResults}
\end{figure*}

\begin{figure}[!t]
\centering
\includegraphics[width=0.5\textwidth]{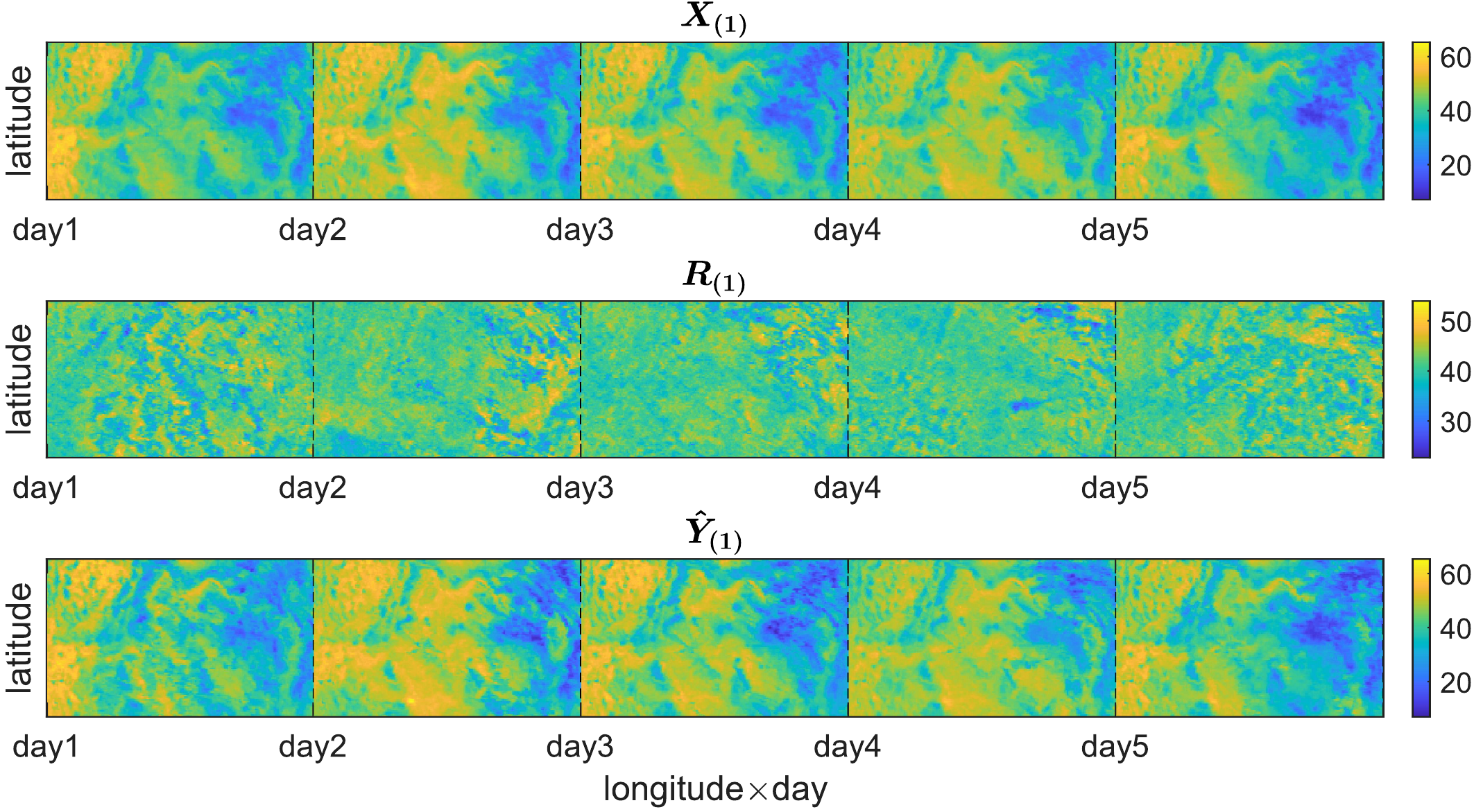}
\caption{Five-day results of BCKL on MODIS 1mon data, including unfolding matrices of the estimated global component $\boldsymbol{X}_{(1)}$, local component $\boldsymbol{R}_{(1)}$, and final output $\hat{\boldsymbol{Y}}_{(1)}$.}
\label{fig_MODIS5day}
\end{figure}

Fig.~\ref{fig_MODIStrain} shows an example of the full and training data in one day. The corresponding estimation surfaces for the data of the same day obtained by different methods  trained on the 3mon data are compared in Fig.~\ref{fig_MODISResults}. As can be seen, BCKL provides the lowest absolute error in most regions of the test data, along with a proper range for the uncertainty. The variances of the uncertainties for BPTF and BKTF are smaller than BCKL, but they miscalculate the temperature values. As a result, most of the true points are not covered by the 95\% intervals, leading to a low CVG value. Additionally, it is clear that the tapering model fails to impute detailed values in the blocked missing areas. Fig.~\ref{fig_MODIS5day} illustrates BCKL estimation details for the first five days of 1mon data. Clearly, the low-rank modeled $\boldsymbol{\mathcal{X}}$ shows consistent global structured data with long-range spatial correlation and daily temporal dependency, while the local GP process $\boldsymbol{\mathcal{R}}$ depicts a detailed image with narrow spatial and temporal correlations; and the combined output approximation $\hat{\boldsymbol{\mathcal{Y}}}$ captures both global and local correlations.

Note that if we represent the LST data on each day as a vector of size $20000$, and concatenate the data of different days into a $\textit{location}\times\textit{day}$, the underlying problem becomes co-kriging for a multivariate spatial process. Although it is tempting to use LMC, the bottleneck is that the $P\times P$ ($P=92$) covariance matrix becomes too large to estimate, not to mention in the additive setting. Instead, we model the daily data as a $100\times 200$ matrix and use low-rank tensor factorization to model/approximate the underlying mean structure for the large 3mon data, which circumvents the curse of dimensionality in traditional geostatistical models.

\subsection{Color image inpainting}

\begin{figure*}[!t]
\centering
\subfigure[BPTF vs. BKTF vs. BCKL.]{
\includegraphics[width=0.335\textwidth]{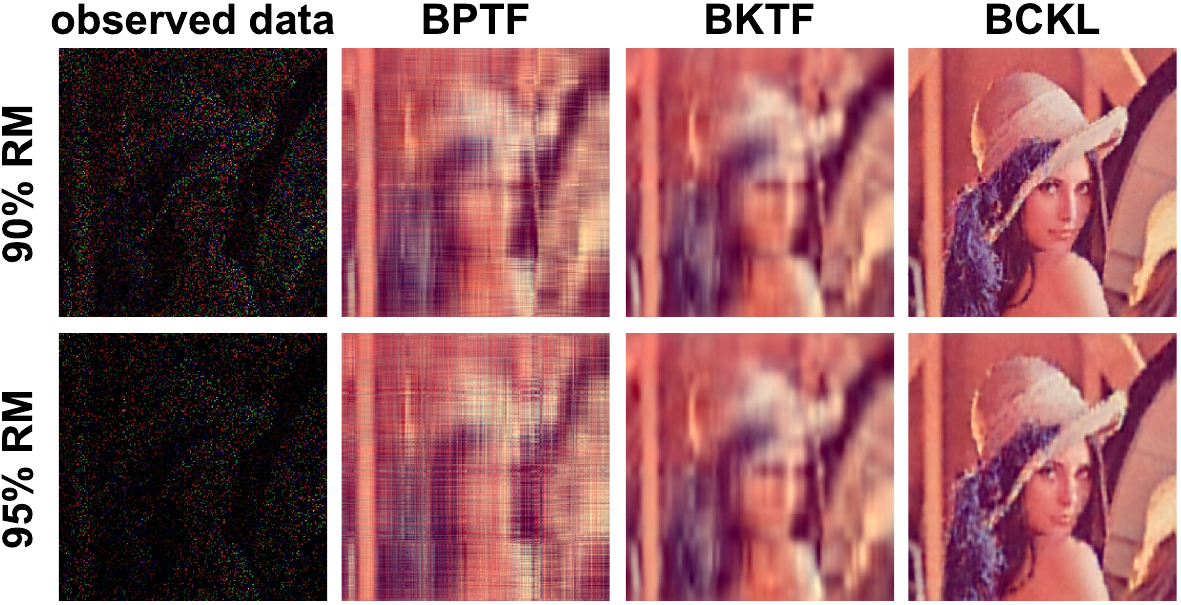}}
\subfigure[BCKL estimated components.]{
\includegraphics[width=0.32\textwidth]{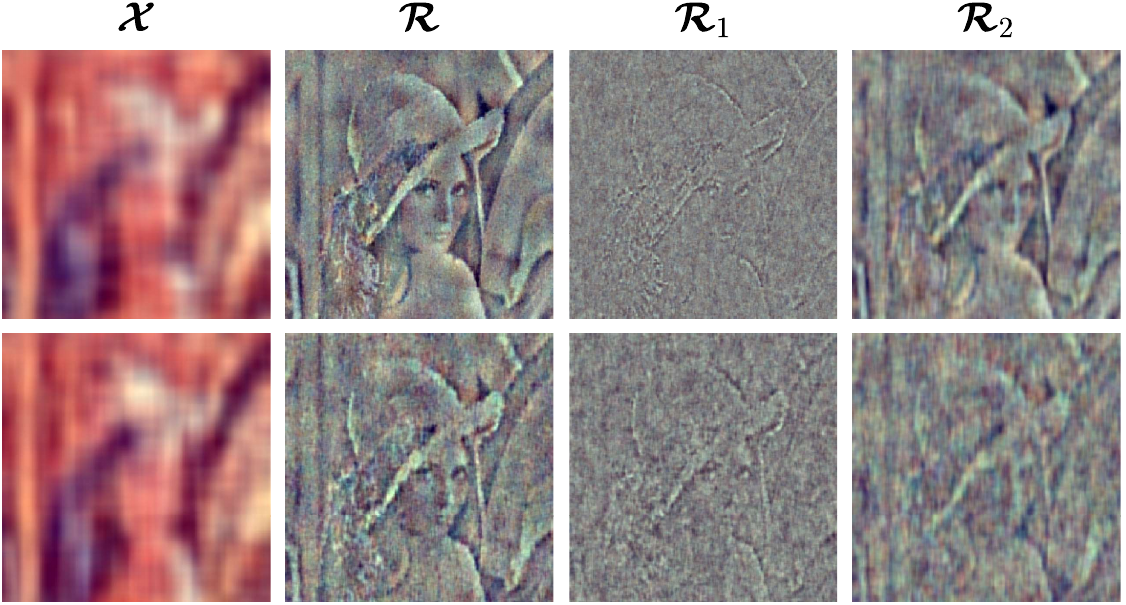}}
\subfigure[Trace plots of BCKL hyperparameters.]{
\includegraphics[width=0.31\textwidth]{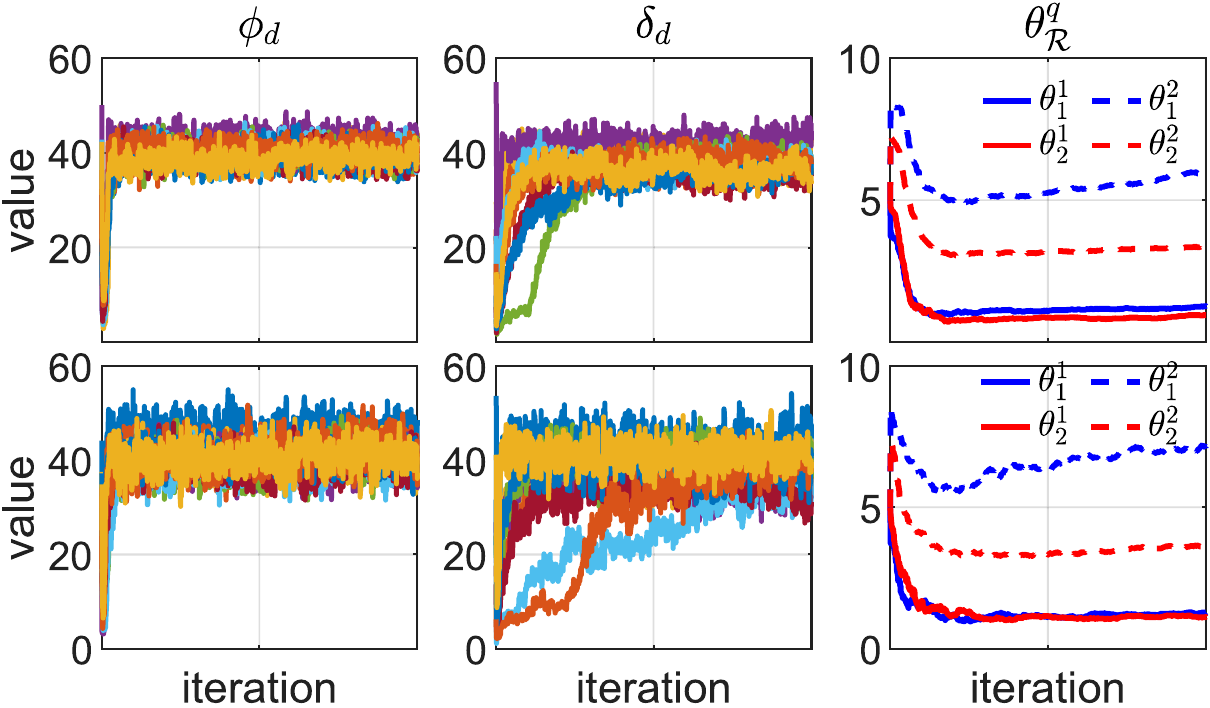}}
\caption{Inpainting results when rank $D=10$. The first and second row of each panel is obtained under 90\% RM and 95\% RM, respectively. $\boldsymbol{\mathcal{R}}_1$ and $\boldsymbol{\mathcal{R}}_2$ in (b) shows the two local components, i.e., $\boldsymbol{r}_1$ and $\boldsymbol{r}_2$. Panel (c) shows 1000 MCMC samples.}
\label{fig_ImageResults}
\end{figure*}

\begin{figure}[!t]
\centering
\includegraphics[width=0.49\textwidth]{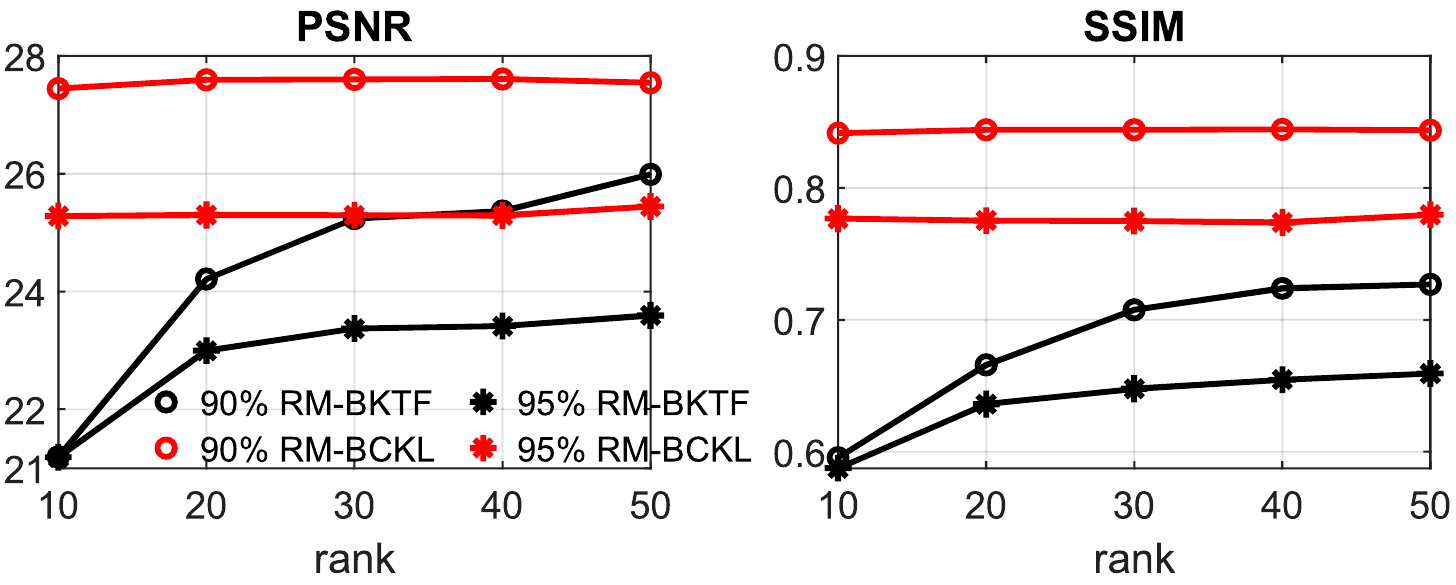}
\caption{Effect of rank $D$ for image inpainting.}
\label{fig_ImageRank}
\end{figure}

\subsubsection{Experimental settings}

We finally evaluate the performance of BCKL on an image inpainting task on \textit{Lena}, which is represented by a $\textit{spatial }\times\textit{ spatial }\times\textit{ channel}$ tensor of size $256\times 256\times 3$. We use Matern 3/2 kernel for $\boldsymbol{u}_{d}$ and $\boldsymbol{v}_{d}$, a Bohman taper with the tapering range being $\{30, 30\}$ as $k_{taper}$, and the same method as for the traffic and MODIS datasets to build $\boldsymbol{K}_{1}^{q}$ and $\boldsymbol{K}_{2}^{q}$ for the local component $\boldsymbol{\mathcal{R}}$. Given that the pixels at different channels are highly correlated, 
we model the covariance matrix for the third dimension of $\boldsymbol{\mathcal{R}}$, i.e., $\boldsymbol{K}_{3}$, as an inverse Wishart distribution of $3\times 3$. We consider uniformly 90\% and 95\% pixels random missing. In both missing scenarios, we experiment with different rank values to determine the effect of rank setting on performance. The value of $Q$ is set as 2, and we take 600 MCMC runs as burn-in and 400 samples for estimation.

\subsubsection{Results}

\begin{table}[!t]
\footnotesize
  \centering
  \caption{Inpainting performance on Lena with 90\% and 95\% random missing (rank $D=10$ and $50$).}
    \begin{tabular}{ll|rrrrrr}
    \toprule
    \multirow{2}{*}{RM} & \multirow{2}{*}{Metrics} & \multicolumn{2}{c}{BPTF} & \multicolumn{2}{c}{BKTF} & \multicolumn{2}{c}{BCKL}  \\
    \cmidrule(r){3-8}
    & & $10$ & $50$ & $10$ & $50$ & $10$ & $50$  \\
    \midrule
    \multirow{5}{*}{90\%} & PSNR & 19.60 & 21.67 & 21.19 & 25.99 & 27.45 & 27.56 \\
    & SSIM & 0.39 & 0.44 & 0.60 & 0.73 & 0.84 & 0.84 \\
    & CRPS & 14.15 & 11.86 & 11.36 & 6.77 & 5.30 & 5.22 \\
    & INT & 59.36 & 57.93 & 43.74 & 35.51 & 12.80 & 12.74 \\
    & CVG & 0.92 & 0.92 & 0.94 & 0.93 & 0.94 & 0.94 \\
    \midrule
    \multirow{5}{*}{95\%} & PSNR & 18.31 & 19.00 & 21.18 & 23.59 & 25.28 & 25.45 \\
    & SSIM & 0.28 & 0.28 & 0.59 & 0.66 & 0.78 & 0.78 \\
    & CRPS & 17.20 & 16.41 & 11.59 & 8.39 & 6.76 & 6.66  \\
    & INT & 71.78 & 84.71 & 40.63 & 42.08 & 13.22 & 13.39   \\
    & CVG & 0.92 & 0.90 & 0.94 & 0.93 & 0.94 & 0.94   \\
    \bottomrule
    \end{tabular}
  \label{Lena-comTable}
\end{table}

Table~\ref{Lena-comTable} lists the quantitative inpainting performance, including PSNR (peak signal-to-noise ratio), SSIM (structural similarity), and uncertainty metrics, of the three Bayesian tensor models, BPTF, BKTF and BCKL, for $D=10$ and $D=50$. The kernel assumptions of BKTF are the same as the settings of the global component in BCKL. We observe that BCKL clearly offers the best recovery performance. The PSNR/SSIM of BCKL with $D=10$ is even better than the results of BKTF with $D=50$, suggesting that the low-rank mean term can substantially enhance the learning of the local component. Fig.~\ref{fig_ImageResults}(a) compares the recovered images of different models when $D=10$. Firstly, BKTF outperforms BPTF, which confirms the importance of smoothness constraints for image inpainting. Secondly, BCKL obviously gives better results than BKTF with more distinct outlines and more detailed contents. Since in this case the low-rank model only captures the global structure and misses the small-scale rough variations, i.e., the residual of BKTF is still highly correlated when the rank is small; while in BCKL, the local component complements the low-rank component and explains the short-range local variations around edges. The global and local components $\left\{\boldsymbol{\mathcal{X}},\boldsymbol{\mathcal{R}},\boldsymbol{r}_1,\boldsymbol{r}_2\text{ (denoted by } \boldsymbol{\mathcal{R}}_1 \text{ and } \boldsymbol{\mathcal{R}}_2\text{)}\right\}$ and kernel length-scales $\left\{\phi_d,\delta_d,\theta_1^q,\theta_2^q\right\}$ learned by BCKL ($D=10$) in both missing scenarios are illustrated in Fig.~\ref{fig_ImageResults}(b) and (c), respectively. We see that the global low-rank component estimates an underlying smooth mean structure; the local GP component, on the other hand, accurately captures the edge information, which is clearly difficult to model by a low-rank factorization. More specifically, the two local components describe edges in different levels: one is finer, see $\boldsymbol{\mathcal{R}}_1$, while one with coarse grained textures, i.e., $\boldsymbol{\mathcal{R}}_2$. The trace plots show an efficient MCMC inference process of BCKL where hundreds of MCMC samples yield converged kernel hyperparameters, and the learned length-scales $\{\phi_d,\delta_d\}$, $\{\theta_1^1,\theta_2^1\}$, $\{\theta_1^2,\theta_2^2\}$ interpretably explain the results of $\boldsymbol{\mathcal{X}}$, $\boldsymbol{\mathcal{R}}_1$, and $\boldsymbol{\mathcal{R}}_2$, respectively. Fig.~\ref{fig_ImageRank} summarizes the two models' sensitivity to rank selection under 90\% and 95\% RM. As we can see, when increasing the rank, BKTF can further capture small-scale variations and thus its performance becomes close to that of BCKL; BCKL offers a superior and robust solution in which the performance is almost invariant to the selection of rank.

\section{Conclusion}\label{sec:conclusion}

In this paper, we propose a novel  Bayesian Complementary Kernelized Learning (BCKL) framework for modeling multidimensional spatiotemporal data. By combining kernelized low-rank factorization with local spatiotemporal GPs, BCKL provides a new probabilistic matrix/tensor factorization scheme that accounts for spatiotemporally correlated residuals. The global long-range structures of the data are effectively captured by the low-rank factorization, while the remaining local short-scale dependencies are characterized by a nonseparable and sparse covariance matrix based on covariance tapering. BCKL is fully Bayesian and can efficiently learn nonstationary and nonseparable multidimensional processes with reliable uncertainty estimates. According to Eq.~\eqref{Eq_full_kernel}, we can see the modeling properties of BCKL: (i) the mean component $\boldsymbol{\mathcal{X}}$ parameterized by kernelized tensor factorization can be viewed as a special case of multidimensional Karhunen–Lo\`{e}ve expansion (or functional PCA) \cite{wang2008karhunen}, which is a powerful tool for generating nonstationary and nonseparable stochastic processes with a low-rank structure; (ii) $\boldsymbol{K}_{\mathcal{R}}$ is {stationary} but {nonseparable}, which is similar to additive kernels/sum-product of separable kernels; (iii) the final represented of BCKL benefits from the computational efficiency inherited from both the low-rank structure of $\boldsymbol{\mathcal{X}}$ and the sparse covariance of residual $\boldsymbol{\mathcal{R}}$, thus providing a highly flexible and effective framework to model complex spatiotemporal data. Numerical experiments on synthetic and real-world datasets demonstrate that BCKL outperforms other baseline models. Moreover, our model can be extended to higher-order tensors, such as MODIS/traffic data with more variables (in addition to temperature/speed) included.


Several directions can be explored for future research. For the global component, instead of CP factorization, the global mean structure can be characterized by the more general Tucker decomposition or tensor-train decomposition. For the latent factor matrix, we can impose additional orthogonal constraints to enhance model identifiability \cite{jauch2021monte,matuk2022bayesian}. In addition, sparse approximations \cite{quinonero2005unifying, banerjee2008gaussian,ren2013hierarchical,luttinen2009variational} can be integrated when the size of each dimension (e.g., $M$ and $T$) becomes large. The local component, on the other hand, can also be estimated by more flexible modeling approaches such as the nearest neighbor GP (NNGP) \cite{datta2016hierarchical} and Gaussian Markov random field (GMRF) \cite{rue2005gaussian}. In terms of applications, this complementary kernel learning framework can also be applied to other completion problems where both long-range patterns and local variations exist, such as graph-regularized collaborative filtering applied in recommendation systems. Lastly, our model can be extended beyond the  Cartesian grid to the more general continuous and misaligned space, such as the work in \cite{schmidt2009function,wilson2015kernel}. 

\ifCLASSOPTIONcompsoc
  \section*{Acknowledgments}
\else
  \section*{Acknowledgment}
\fi


This work was supported in part by the Natural Sciences and Engineering Research Council (NSERC) of Canada Discovery Grant RGPIN-2019-05950, in part by the Fonds de recherche du Qu\'{e}bec--Nature et technologies
(FRQNT) Research Support for New Academics \#283653, and in part by the Canada Foundation for Innovation (CFI) John R. Evans Leaders Fund (JELF). M. Lei would like to thank the Institute for Data Valorization (IVADO) for providing the Excellence Ph.D. Scholarship.

\bibliographystyle{IEEEtran}

\bibliography{ref}

\end{document}